\documentclass[10pt,journal,compsoc]{IEEEtran}

\ifCLASSOPTIONcompsoc
  \usepackage{cite}
\else
  \usepackage{cite}
\fi

\ifCLASSINFOpdf
  \usepackage[pdftex]{graphicx}
\else
  \usepackage[dvips]{graphicx}
\fi

\usepackage{amsmath}
\usepackage{amssymb}
\usepackage{amsfonts} 

\ifCLASSOPTIONcompsoc
  \usepackage[caption=false,font=footnotesize,labelfont=sf,textfont=sf]{subfig}
\else
  \usepackage[caption=false,font=footnotesize]{subfig}
\fi

\usepackage{hyperref}
\usepackage{url}
\usepackage{color}
\usepackage{colortbl}
\usepackage{soul}
\usepackage{multirow}

\usepackage{rotating,siunitx} %
\usepackage{tabularx}
\usepackage{hyphenat}
\usepackage{longtable}

\usepackage{float}

\usepackage{etoolbox}
\makeatletter
\patchcmd{\@makecaption}
  {\scshape}
  {}
  {}
  {}
\makeatother

\usepackage{algorithm}      % algorithm
\usepackage[noend]{algpseudocode} % algorithm

\newcommand{\algrule}[1][.2pt]{\par\vskip.1\baselineskip\hrule height #1\par\vskip.1\baselineskip}

\usepackage{pifont}% http://ctan.org/pkg/pifont

\newcommand{\etal}{\textit{et al}.}
\newcommand{\ie}{\textit{i}.\textit{e}.}
\newcommand{\eg}{\textit{e}.\textit{g}.}

\definecolor{LightRed}{rgb}{1,0.92,0.92}
\definecolor{LightOrange}{rgb}{1,0.95,0.88}
\definecolor{LightYellow}{rgb}{1.0,1.0,0.84}
\definecolor{LightGreen}{rgb}{0.9,1.0,0.88}
\definecolor{LightCyan}{rgb}{0.9,1,1}
\definecolor{LightBlue}{rgb}{0.9,0.94,1}
\definecolor{LightIndigo}{rgb}{0.92,0.9,1}
\definecolor{LightMagenta}{rgb}{0.96,0.86,1}
\definecolor{DirtyWhite}{rgb}{0.96,0.96,0.96}

\begin{document}
\title{Diffusion Models in Vision: A Survey}

\author{Florinel-Alin~Croitoru,
        {Vlad}~Hondru,
        {Radu Tudor}~Ionescu,~\IEEEmembership{Member,~IEEE,}
        and~Mubarak~Shah,~\IEEEmembership{Fellow,~IEEE}
       % and~Marius~Popescu% <-this % stops a space
\IEEEcompsocitemizethanks{\IEEEcompsocthanksitem F.A. Croitoru, V. Hondru and R.T. Ionescu are with the Department of Computer Science, University of Bucharest, Bucharest, Romania. F.A. Croitoru and V. Hondru have contributed equally. R.T. Ionescu is the corresponding author.\protect\\
E-mail: raducu.ionescu@gmail.com
\IEEEcompsocthanksitem M. Shah is with the Center for Research in Computer Vision (CRCV), Department of Computer Science, University of Central Florida, Orlando, FL, 32816.}
% <-this % stops an unwanted space
\thanks{Manuscript received April 19, 2022; revised August 26, 2022.}}


\markboth{IEEE Transactions on Pattern Analysis and Machine Intelligence,~Vol.~14, No.~8, August~2022}%
{Croitoru \MakeLowercase{\textit{et al.}}: Diffusion Models: A Survey}


\IEEEtitleabstractindextext{%
\begin{abstract}
Denoising diffusion models represent a recent emerging topic in computer vision, demonstrating remarkable results in the area of generative modeling. A diffusion model is a deep generative model that is based on two stages, a forward diffusion stage and a reverse diffusion stage. In the forward diffusion stage, the input data is gradually perturbed over several steps by adding Gaussian noise. In the reverse stage, a model is tasked at recovering the original input data by learning to gradually reverse the diffusion process, step by step. Diffusion models are widely appreciated for the quality and diversity of the generated samples, despite their known computational burdens, \ie~low speeds due to the high number of steps involved during sampling. In this survey, we provide a comprehensive review of articles on denoising diffusion models applied in vision, comprising both theoretical and practical contributions in the field. First, we identify and present three generic diffusion modeling frameworks, which are based on denoising diffusion probabilistic models, noise conditioned score networks, and stochastic differential equations. We further discuss the relations between diffusion models and other deep generative models, including variational auto-encoders, generative adversarial networks, energy-based models, autoregressive models and normalizing flows. Then, we introduce a multi-perspective categorization of diffusion models applied in computer vision. Finally, we illustrate the current limitations of diffusion models and envision some interesting directions for future research.
\end{abstract}
}

\maketitle

\setlength{\abovedisplayskip}{3.5pt}
\setlength{\belowdisplayskip}{3.5pt}

\IEEEdisplaynontitleabstractindextext

\IEEEpeerreviewmaketitle

\IEEEraisesectionheading{\section{Introduction}\label{sec:introduction}}

\begin{figure}[!t]
\begin{center}
\centerline{\includegraphics[width=0.85\linewidth]{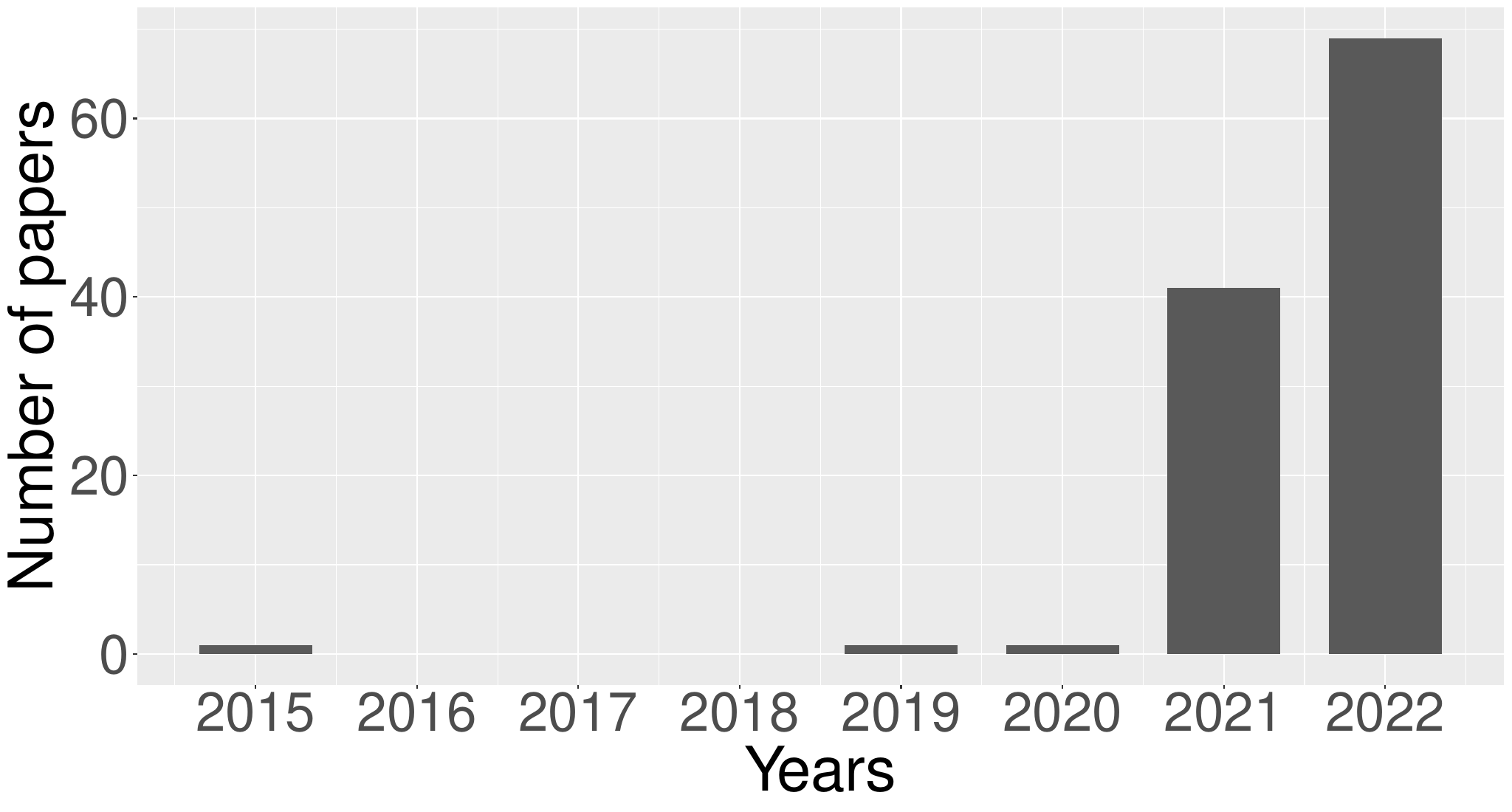}}
\vspace{-0.25cm}
\caption{The rough number of papers on diffusion models per year.}
\label{fig_stats}
\vspace{-0.9cm}
\end{center}
\end{figure}

\begin{figure*}[!t]
\begin{center}
\centerline{\includegraphics[width=1.0\linewidth]{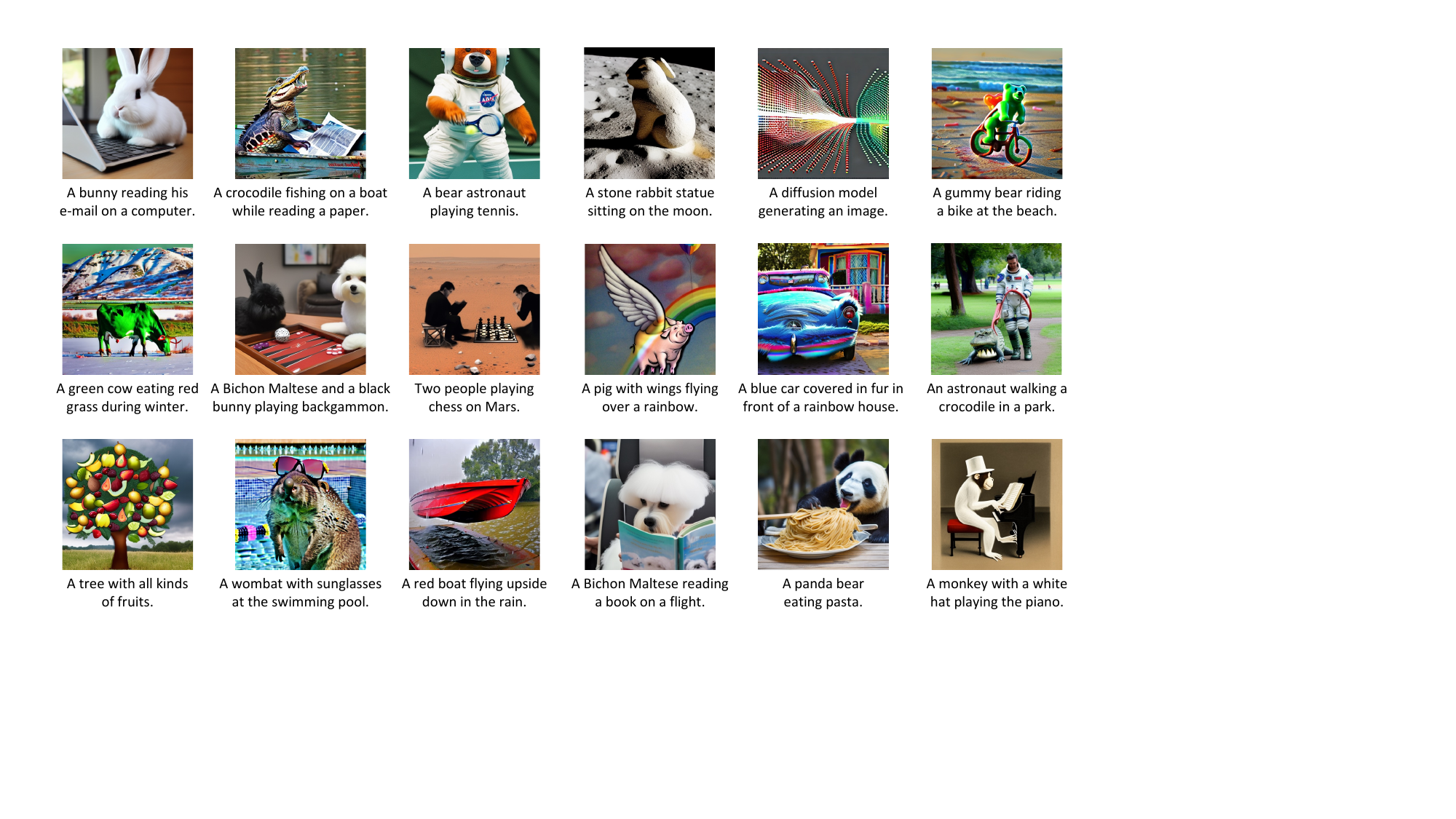}}
\vspace{-0.25cm}
\caption{Images generated by Stable Diffusion \cite{rombach-CVPR-2022} based on various text prompts, via the \url{https://beta.dreamstudio.ai/dream} platform.}
\label{fig_tti_examples}
\vspace{-0.4cm}
\end{center}
\end{figure*}

\IEEEPARstart{D}{iffusion} models \cite{sohl-icml-2015, ho-NeurIPS-2020, song-NeurIPS-2019, song-ICLR-2021, dhariwal-NeurIPS-2021, nichol-ICML-2021, song-ICLR-2021b, watson-ICLR-2021, shi-arXiv-2022b, rombach-CVPR-2022, rombach-arXiv-2022} form a category of deep generative models which has recently become one of the hottest topics in computer vision (see Figure~\ref{fig_stats}), showcasing impressive generative capabilities, ranging from the high level of details to the diversity of the generated examples. We can even go as far as stating that these generative models raised the bar to a new level in the area of generative modeling, particularly referring to models such as Imagen \cite{saharia-arXiv-2022} and Latent Diffusion Models (LDMs) \cite{rombach-CVPR-2022}. This statement is confirmed by the image samples illustrated in Figure~\ref{fig_tti_examples}, which are generated by Stable Diffusion, a version of LDMs \cite{rombach-CVPR-2022} that generates images based on text prompts. The generated images exhibit very few artifacts and are very well aligned with the text prompts. Notably, the prompts are purposely chosen to represent unrealistic scenarios (never seen at training time), thus demonstrating the high generalization capacity of diffusion models.

To date, diffusion models have been applied to a wide variety of generative modeling tasks, such as image generation \cite{sohl-icml-2015, song-NeurIPS-2019, ho-NeurIPS-2020, song-NeurIPS-2020, song-ICLR-2021, dhariwal-NeurIPS-2021, nichol-arXiv-2021, song-ICLR-2021b, song-NeurIPS-2021, sinha-NeurIPS-2021, vahdat-NeurIPS-2021, saharia-arXiv-2021, nichol-ICML-2021, pandey-NeurIPSW-2021, rombach-CVPR-2022, bao-arXiv-2022, dockhorn-arXiv-2021, rombach-arXiv-2022, liu-arXiv-2022b, jiang-arXiv-2022}, image super-resolution \cite{saharia-arXiv-2021, batzolis-arXiv-2021, daniels-NeurIPS-2021, rombach-CVPR-2022, chung-CVPR-2022, kawar-arXiv-2022}, image inpainting \cite{sohl-icml-2015, song-NeurIPS-2019, song-ICLR-2021, esser-NeurIPS-2021, batzolis-arXiv-2021, lugmayr-CVPR-2022, rombach-CVPR-2022, chung-CVPR-2022, jing-arXiv-2022}, image editing \cite{avrahami-CVPR-2022, choi-arXiv-2021, meng-arXiv-2021}, image-to-image translation \cite{saharia-SIGGRAPH-2022, choi-arXiv-2021, zhao-arXiv-2022, wang-arXiv-2022c, li-arXiv-2022, wolleb-arXiv-2022b}, among others. Moreover, the latent representation learned by diffusion models was also found to be useful in discriminative tasks, \eg~image segmentation \cite{baranchuk-arXiv-2021, graikos-arXiv-2022, wolleb-arXiv-2021, amit-arXiv-2021}, classification \cite{zimmermann-arXiv-2021} and anomaly detection \cite{pinaya-arXiv-2022, wolleb-arXiv-2022a, Wyatt-CVPRW-2022}. This confirms the broad applicability of denoising diffusion models, indicating that further applications are yet to be discovered. Additionally, the ability to learn strong latent representations creates a connection to representation learning \cite{Bengio-TPAMI-2013, Goodfellow-MIT-2016}, a comprehensive domain that studies ways to learn powerful data representations, covering multiple approaches ranging from the design of novel neural architectures \cite{Hinton-SC-2006,Kingma-ICLR-2014b, Higgins-ICLR-2017,Goodfellow-NIPS-2014} to the development of learning strategies \cite{Caron-NIPS-2020, Chen-ICML-2020,Croitoru-CVPRW-2022,Oord-arXiv-2018,Samuli-ICLR-2017, Tarvainen-NIPS-2017}. 

According to the graph shown in Figure~\ref{fig_stats}, the number of papers on diffusion models is growing at a very fast pace.
To outline the past and current achievements of this rapidly developing topic, we present a comprehensive review of articles on denoising diffusion models in computer vision. More precisely, we survey articles that fall in the category of generative models defined below. \emph{Diffusion models} represent a category of deep generative models that are based on $(i)$ a forward diffusion stage, in which the input data is gradually perturbed over several steps by adding Gaussian noise, and $(ii)$ a reverse (backward) diffusion stage, in which a generative model is tasked at recovering the original input data from the diffused (noisy) data by learning to gradually reverse the diffusion process, step by step. 

\begin{figure*}[!th]
\begin{center}
\centerline{\includegraphics[width=0.82\linewidth]{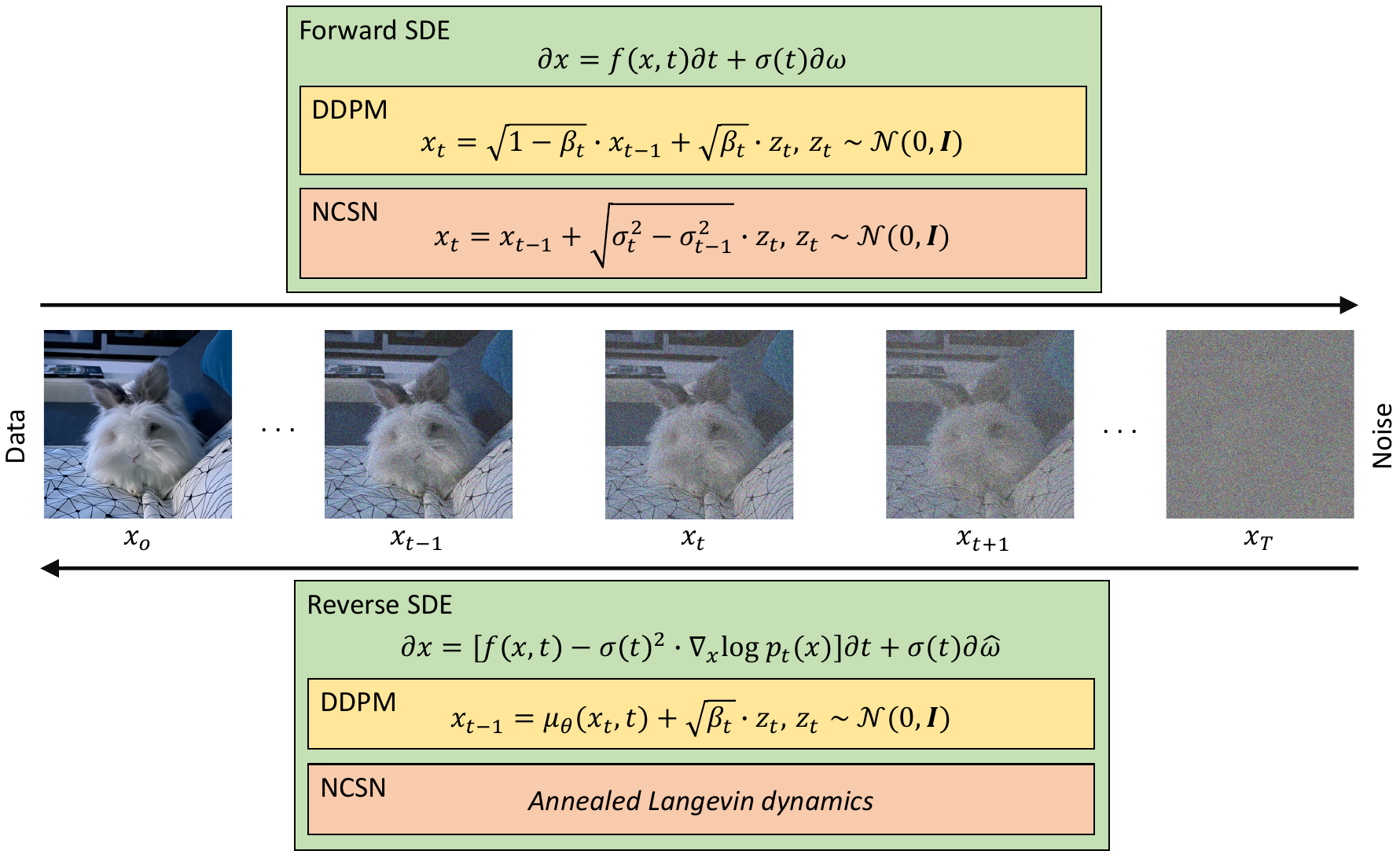}}
\vspace{-0.25cm}
\caption{A generic framework composing three alternative formulations of diffusion models based on: stochastic differential equations (SDEs), denoising diffusion probabilistic models (DDPMs) and noise conditioned score networks (NCSNs). In general, a diffusion model consists of two processes. The first one, called the forward process, transforms data into noise, while the second one is a generative process that reverses the effect of the forward process. This latter process learns to transform the noise back into data. We illustrate these processes for all three formulations. The forward SDE shows that a change over time in $x$ is modeled by a function $f$ plus a stochastic component $\partial \omega \sim \mathcal{N}(0,\partial t)$ scaled by $\sigma(t)$. We underline that different choices of $f$ and $\sigma$ will lead to different diffusion processes. This is why the SDE formulation is a generalization of the other two. The reverse (generative) SDE shows how to change $x$ in order to recover the data from pure noise. We keep the random component and modify the deterministic one using the gradients of the log probability $\nabla_x \log{p_t(x)}$, so that $x$ moves to regions where the data density $p(x)$ is high. DDPMs sample the data points during the forward process from a normal distribution $\mathcal{N}\!\left(x_t;\! \sqrt{1\!-\!\beta_t}\!\cdot\! x_{t-1}, \beta_t\!\cdot\!\mathbf{I}\right)$, where $\beta_t\ll1$. This iterative sampling slowly destroys information in data, and replaces it with Gaussian noise. The sampling is illustrated via the reparametrization trick (see details in Section~\ref{sec2.1}). The reverse process of DDPM also performs iterative sampling from a normal distribution, but the mean $\mu_\theta(x_t,t)$  of the distribution is derived by subtracting the noise, estimated by a neural network, from the image at the previous step $x_t$. The variance is equal to the one used in the forward process. The initial image going into the reverse process contains only Gaussian noise. The forward process of NCSN simply adds normal noise to the image at the previous step. This can also be seen as sampling from a normal distribution $\mathcal{N}(x_t;x_{t-1},(\sigma_t^2-\sigma_{t-1}^2))\cdot \mathbf{I})$, with the mean being the image at the previous step. The reverse process of NCSN is based on an algorithm described in Section~\ref{sec_NCSN}. Best viewed in color.}
\label{fig_pipeline}
\vspace{-0.4cm}
\end{center}
\end{figure*}

We underline that there are at least three sub-categories of diffusion models that comply with the above definition. 
The first sub-category comprises denoising diffusion probabilistic models (DDPMs) \cite{ho-NeurIPS-2020, sohl-icml-2015}, which are inspired by the non-equilibrium thermodynamics theory. DDPMs are latent variable models that employ latent variables to estimate the probability distribution. From this point of view, DDPMs can be viewed as a special kind of variational auto-encoders (VAEs) \cite{Kingma-ICLR-2014b}, where the forward diffusion stage corresponds to the encoding process inside VAE, while the reverse diffusion stage corresponds to the decoding process. The second sub-category is represented by noise conditioned score networks (NCSNs) \cite{song-NeurIPS-2019}, which are based on training a shared neural network via score matching to estimate the score function (defined as the gradient of the log density) of the perturbed data distribution at different noise levels. Stochastic differential equations (SDEs) \cite{song-ICLR-2021} represent an alternative way to model diffusion, forming the third sub-category of diffusion models. Modeling diffusion via forward and reverse SDEs leads to efficient generation strategies as well as strong theoretical results \cite{huang-NeurIPS-2021}. This latter formulation (based on SDEs) can be viewed as a generalization over DDPMs and NCSNs.

We identify several defining design choices and synthesize them into three generic diffusion modeling frameworks corresponding to the three sub-categories introduced above. To put the generic diffusion modeling framework into context, we further discuss the relations between diffusion models and other deep generative models. More specifically, we describe the relations to variational auto-encoders (VAEs) \cite{Kingma-ICLR-2014b}, generative adversarial networks (GANs) \cite{Goodfellow-NIPS-2014}, energy-based models (EBMs) \cite{lecun-PSD-2006, ngiam-ICML-2011}, autoregressive models \cite{Oord-NeurIPS-2016} and normalizing flows \cite{dinh-ICLR-2015, dinh-ICLR-2017}. Then, we introduce a multi-perspective categorization of diffusion models applied in computer vision, classifying the existing models based on several criteria, such as the underlying framework, the target task, or the denoising condition. Finally, we illustrate the current limitations of diffusion models and envision some interesting directions for future research. For example, perhaps one of the most problematic limitations is the poor time efficiency during inference, which is caused by a very high number of evaluation steps, \eg~thousands, to generate a sample \cite{ho-NeurIPS-2020}. Naturally, overcoming this limitation without compromising the quality of the generated samples represents an important direction for future research.

In summary, our contribution is twofold:
\begin{itemize}
    \item Since many contributions based on diffusion models have recently emerged in vision, we provide a comprehensive and timely literature review of denoising diffusion models applied in computer vision, aiming to provide a fast understanding of the generic diffusion modeling framework to our readers.
    \item We devise a multi-perspective categorization of diffusion models, aiming to help other researchers working on diffusion models applied to a specific domain in quickly finding relevant related works in the respective domain.
\end{itemize}

% Proposed a thermal data set for anomaly detection
% The masked convolution is providing consistent improvement across the cross domain. 
\section{Generic Framework}

Diffusion models are a class of probabilistic generative models that learn to reverse a process that gradually degrades the training data structure. Thus, the training procedure involves two phases: the forward diffusion process and the backward denoising process.   

The former phase consists of multiple steps in which low-level noise is added to each input image, where the scale of the noise varies at each step. The training data is progressively destroyed until it results in pure Gaussian noise. 

The latter phase is represented by reversing the forward diffusion process. The same iterative procedure is employed, but backwards: the noise is sequentially removed, and hence, the original image is recreated. Therefore, at inference time, images are generated by gradually reconstructing them starting from random white noise. The noise subtracted at each time step is estimated via a neural network, typically based on a U-Net architecture \cite{Ronneberger-MICCAI-2015}, allowing the preservation of dimensions.

In the following three subsections, we present three formulations of diffusion models, namely denoising diffusion probabilistic models, noise conditioned score networks, and the approach based on stochastic differential equations that generalizes over the first two methods. For each formulation, we describe the process of adding noise to the data, the method which learns to reverse this process, and how new samples are generated at inference time. In Figure~\ref{fig_pipeline}, all three formulations are illustrated as a generic framework.
We dedicate the last subsection to discussing connections to other deep generative models.

\subsection{Denoising Diffusion Probabilistic Models (DDPMs)} \label{sec2.1}

\noindent {\bf Forward process.} DDPMs \cite{sohl-icml-2015, ho-NeurIPS-2020} slowly corrupt the training data using Gaussian noise. Let $p(x_0)$ be the data density, where the index $0$ denotes the fact that the data is uncorrupted (original). Given an uncorrupted training sample $x_0 \sim p(x_0)$, the noised versions $x_1,x_2\dots, x_T$ are obtained according to the following Markovian process:
\begin{equation}
    \label{eq_ddpm_1}
    \!\!p(x_t|x_{t-1})\!=\!\mathcal{N}\!\left(x_t;\! \sqrt{1\!-\!\beta_t}\!\cdot\! x_{t-1}, \beta_t\!\cdot\!\mathbf{I}\right)\!, \forall t\!\in\!\{1,\dots,T\},\!
\end{equation}
where $T$ is the number of diffusion steps, $\beta_1, \dots, \beta_T \in [0,1)$ are hyperparameters representing the variance schedule across diffusion steps, $\mathbf{I}$ is the identity matrix having the same dimensions as the input image $x_0$, and $\mathcal{N}(x; \mu, \sigma)$ represents the normal distribution of mean $\mu$ and covariance $\sigma$ that produces $x$. An important property of this recursive formulation is that it also allows the direct sampling of $x_t$, when $t$ is drawn from a uniform distribution, \ie~$\forall t \sim \mathcal{U}(\{1, \dots, T\})$:
\begin{equation}
 \label{eq_ddpm_2}
    \begin{split}
        p(x_t|x_0) &= \mathcal{N}\!\left(x_t; \sqrt{\hat{\beta_t}}\cdot x_0, (1-\hat{\beta_t})\cdot\mathbf{I}\right), 
    \end{split}
\end{equation}
where $\hat{\beta_t} = \prod_{i=1}^{t}\alpha_i$ and  $\alpha_t=1-\beta_t$. Essentially, Eq.~\eqref{eq_ddpm_2} shows that we can sample any noisy version $x_t$ via a single step, if we have the original image $x_0$ and fix a variance schedule $\beta_t$.

The sampling from $p(x_t|x_0)$ is performed via a reparametrization trick. In general, to standardize a sample $x$ of a normal distribution $x \sim \mathcal{N}(\mu,\sigma^2\cdot\mathbf{I})$, we subtract the mean $\mu$ and divide by the standard deviation $\sigma$, resulting in a sample $z=\frac{x-\mu}{\sigma}$ of the standard normal distribution $z\sim \mathcal{N}(0,\mathbf{I})$. The reparametrization trick does the inverse of this operation, starting with $z$ and yielding the sample $x$ by multiplying $z$ with the standard deviation $\sigma$ and adding the mean $\mu$. If we translate this process to our case, then $x_t$ is sampled from $p(x_t|x_0)$ as follows:
\begin{equation}
 \label{rep_trick_sampling}
    \begin{split}
        x_t &= \sqrt{\hat{\beta_t}}\cdot x_0 + \sqrt{(1-\hat{\beta_t})}\cdot z_t, 
    \end{split}
\end{equation}
where $z_t \sim \mathcal{N}(0,\mathbf{I})$.

\noindent {\bf Properties of $\beta_t$.} If the variance schedule $(\beta_t)_{t=1}^{T}$ is chosen such that $\hat{\beta}_T \rightarrow 0$, then, according to Eq.~\eqref{eq_ddpm_2}, the distribution of $x_T$ should be well approximated by the standard Gaussian distribution $ \pi(x_T) = \mathcal{N}(0,\mathbf{I})$. Moreover, if each $(\beta_t)_{t=1}^{T} \ll 1$, then the reverse steps $p(x_{t-1}|x_t)$ have the same functional form as the forward process $p(x_t|x_{t-1})$ \cite{sohl-icml-2015, feller-SMSP-1949}. Intuitively, the last statement is true when $x_t$ is created with a very small step, as it becomes more likely that $x_{t-1}$ comes from a region close to where $x_t$ is observed, which allows us to model this region with a Gaussian distribution. 
To conform to the aforementioned properties, Ho \etal~\cite{ho-NeurIPS-2020} choose $(\beta_t)_{t=1}^{T}$ to be linearly increasing constants between $\beta_1 = 10^{-4}$ and $\beta_T = 2 \cdot 10^{-2}$, where $T=1000$.
 
\noindent {\bf Reverse process.} By leveraging the above properties, we can generate new samples from $p(x_0)$ if we start from a sample $x_T \sim \mathcal{N}(0,\mathbf{I})$ and follow the reverse steps $p(x_{t-1}|x_t) = \mathcal{N}(x_{t-1};\mu(x_t,t),\Sigma(x_t,t))$. To approximate these steps, we can train a neural network $p_{\theta}(x_{t-1}|x_t) = \mathcal{N}(x_{t-1};\mu_\theta(x_t,t),\Sigma_\theta(x_t,t))$ that receives as input the noisy image $x_t$ and the embedding at time step $t$, and learns to predict the mean $\mu_\theta(x_t,t)$ and the covariance $\Sigma_\theta(x_t,t)$.

In an ideal scenario, we would train the neural network with a maximum likelihood objective such that the probability assigned by the model $p_\theta(x_0)$ to each training example $x_0$ is as large as possible. However, $p_\theta(x_0)$ is intractable because we have to marginalize over all the possible reverse trajectories to compute it. The solution to this problem \cite{sohl-icml-2015, ho-NeurIPS-2020} is to minimize a variational lower-bound of the negative log-likelihood instead, which has the following formulation:
\begin{equation}
 \label{eq_ddpm_3}
    \begin{split}
        \mathcal{L}_{\scriptsize{\mbox{\emph{vlb}}}} = &-\log p_\theta(x_0|x_1)
        + \mbox{\emph{KL}}\left(p(x_T|x_0) \Vert \pi(x_T)\right)\\
        &+ \sum_{t > 1} \mbox{\emph{KL}}(p(x_{t-1}|x_t,x_0)\Vert p_\theta(x_{t-1}|x_t)),
    \end{split}
\end{equation}
where $\mbox{\emph{KL}}$ denotes the Kullback-Leibler divergence between two probability distributions. The full derivation of this objective is presented in Appendix~\ref{appendix_vlb}. Upon analyzing each component, we can see that the second term can be removed because it does not depend on $\theta$. The last term shows that the neural network is trained such that, at each time step $t$, $p_\theta(x_{t-1}|x_t)$ is as close as possible to the true posterior of the forward process when conditioned on the original image. Moreover, it can be proven that the posterior $p(x_{t-1}|x_t,x_0)$ is a Gaussian distribution, implying closed-form expressions for the $\mbox{\emph{KL}}$ divergences. 

Ho \etal~\cite{ho-NeurIPS-2020} propose to fix the covariance $\Sigma_\theta(x_t,t)$ to a constant value and rewrite the mean $\mu_\theta(x_t,t)$ as a function of noise, as follows:
\begin{equation}\label{eq_ddpm_miu}
    \mu_\theta=\frac{1}{\sqrt{\alpha_t}}\cdot\left(x_t - \frac{1-\alpha_t}{\sqrt{1-\hat{\beta_t}}} \cdot z_\theta(x_t,t)\right)\!.
\end{equation}
These simplifications (more details in Appendix~\ref{noise_estimation}) unlocked a new formulation of the objective $\mathcal{L}_{\scriptsize{\mbox{\emph{vlb}}}}$, which measures, for a random time step $t$ of the forward process, the distance between the real noise $z_t$ and the noise estimation $z_\theta(x_t,t)$ of the model:
\begin{equation}
    \label{eq_ddpm_4}
    \mathcal{L}_{\scriptsize{\mbox{\emph{simple}}}} = \mathbb{E}_{t\sim[1,T]}\mathbb{E}_{x_0 \sim p(x_0)}\mathbb{E}_{z_t \sim \mathcal{N}(0,\mathbf{I})}\!\left\lVert z_t-z_\theta(x_t,t)\right\rVert^{2},
\end{equation}
where $\mathbb{E}$ is the expected value, and $z_\theta(x_t,t)$ is the network predicting the noise in $x_t$. We underline that $x_t$ is sampled via Eq.~\eqref{rep_trick_sampling}, where we use a random image $x_0$ from the training set.

The generative process is still defined by $p_\theta(x_{t-1}|x_t)$, but the neural network does not predict the mean and the covariance directly. Instead, it is trained to predict the noise from the image, and the mean is determined according to Eq.~\eqref{eq_ddpm_miu}, while the covariance is fixed to a constant. Algorithm \ref{alg_sampling_ddpm} formalizes the whole generative procedure.

\begin{algorithm}[t]
\caption{DDPM sampling method\label{alg_sampling_ddpm}}
\textbf{Input}: 

$T$ -- the number of diffusion steps.

$\sigma_1, \dots, \sigma_T$ -- the standard deviations for the reverse transitions.
\vspace{0.2em}
\algrule
\vspace{0.2em}
\textbf{Output}: 

$x_0$ -- the sampled image.
\vspace{0.2em}
\algrule
\vspace{0.2em}
\textbf{Computation}:
\begin{algorithmic}[1]
\State $x_T \sim \mathcal{N}(0, \mathbf{I})$
\For {$t = T, \dots, 1$}
    \If {$t > 1$}
    \State $z \sim \mathcal{N}(0, \mathbf{I})$
    \Else
    \State $z = 0$
    \EndIf
    \State $\mu_\theta = \frac{1}{\sqrt{\alpha_t}}\cdot\left(x_t - \frac{1-\alpha_t}{\sqrt{1-\hat{\beta_t}}} \cdot z_\theta(x_t,t)\right)\!$
    \State $x_{t-1} = \mu_\theta + \sigma_t \cdot z$
\EndFor
\end{algorithmic}
\end{algorithm}

\subsection{Noise Conditioned Score Networks (NCSNs)}
\label{sec_NCSN}

The score function of some data density $p(x)$ is defined as the gradient of the log density with respect to the input, $\nabla_x \log{p(x)}$. The directions given by these gradients are used by the Langevin dynamics algorithm \cite{song-NeurIPS-2019} to move from a random sample ($x_0$) towards samples ($x_N$) in regions with high density. \emph{Langevin dynamics} is an iterative method inspired from physics that can be used for data sampling. In physics, this method is used to determine the trajectory of a particle in a molecular system that allows interactions between the particle and the other molecules. The trajectory of the particle is influenced by a drag force of the system and by a random force motivated by the fast interactions between the molecules. In our case, we can think of the gradient of the log density as a force that drags a random sample through the data space into regions with high data density $p(x)$. There is another term $\omega_i$ that accounts, in physics, for the random force, but for us, it is useful to escape local minima. Lastly, a value denoted by $\gamma$ weighs the impact of both forces, because it represents the friction coefficient of the environment where the particle resides. From the sampling point of view, $\gamma$ controls the magnitude of the updates. In summary, the iterative updates of the Langevin dynamics are the following:
\begin{equation}
    \label{eq_ncsn_1}
    x_i = x_{i-1} + \frac{\gamma}{2} \nabla_x \log{p(x)} + \sqrt{\gamma}\cdot \omega_i,
\end{equation}
where $i \in \{1,\dots, N\}$, $\gamma$ controls the magnitude of the update in the direction of the score, $x_0$ is sampled from a prior distribution, the noise $\omega_i \sim \mathcal{N}(0,\mathbf{I})$ addresses the issue of getting stuck in local minima, and the method is applied recursively for $N \rightarrow \infty $ steps.
Therefore, a generative model can employ the above method to sample from $p(x)$ after estimating the score with a neural network $s_\theta(x) \approx \nabla_x \log{p(x)}$. This network can be trained via score matching, a method that requires the optimization of the following objective:
\begin{equation}
    \label{eq_ncsn_2}
    \mathcal{L}_{\scriptsize{\mbox{\emph{sm}}}} = \mathbb{E}_{x \sim p(x)}\left\lVert s_\theta(x) - \nabla_x \log {p(x)}\right\rVert_2^2.
\end{equation}
In practice, it is impossible to minimize this objective directly, because $\nabla_x \log {p(x)}$ is unknown. However, there are other methods such as denoising score matching \cite{Vincent-NC-2011} and sliced score matching \cite{Song-UAI-2019} that overcome this problem.

Although the described approach can be used for data generation, Song \etal~\cite{song-NeurIPS-2019} emphasize several issues when applying this method on real data. Most of the problems are linked with the manifold hypothesis. For example, the score estimation $s_\theta(x)$ is inconsistent when the data resides on a low-dimensional manifold and, among other implications, this could cause the Langevin dynamics to never converge to the high-density regions. In the same work \cite{song-NeurIPS-2019}, the authors demonstrate that these problems can be addressed by perturbing the data with Gaussian noise at different scales. Furthermore, they propose to learn score estimates for the resulting noisy distributions via a single noise conditioned score network (NCSN). Regarding the sampling, they adapt the strategy in Eq.~\eqref{eq_ncsn_1} and use the score estimates associated with each noise scale.

Formally, given a sequence of Gaussian noise scales $\sigma_1 < \sigma_2 < \dots < \sigma_T$ such that $p_{\sigma_1}(x) \approx p(x_0)$ and $p_{\sigma_T}(x) \approx \mathcal{N}(0, \mathbf{I})$, we can train an NCSN $s_\theta(x,\sigma_t)$ with denoising score matching so that $s_\theta(x,\sigma_t) \approx \nabla_x \log(p_{\sigma_t}(x))$, $\forall t \in \{1,\dots,T \}$. We can derive $\nabla_x \log(p_{\sigma_t}(x))$ as follows:
\begin{equation}
    \label{eq_ncsn_3}
    \begin{split}
        \nabla_{x_t} \log{p_{\sigma_t}(x_t|x)} &= -\frac{x_t-x}{\sigma_t^2},
    \end{split}
\end{equation}
given that:
\begin{equation}
    \label{eq_ncsn_33}
    \begin{split}
        p_{\sigma_t}(x_t|x) &= \mathcal{N}(x_t; x, \sigma_t^2\cdot\mathbf{I}) \\
        &= \frac{1}{\sigma_t\cdot \sqrt{2\pi}} \cdot \mbox{\emph{exp}}\!\left({{-\frac{1}{2}\cdot\left(\frac{x_t-x}{\sigma_t}\right)^2}}\right)\!,
    \end{split}
\end{equation}
where $x_t$ is a noised version of $x$, and $\mbox{\emph{exp}}$ is the exponential function.
Consequently, generalizing Eq.~\eqref{eq_ncsn_2} for all $(\sigma_t)_{t=1}^{T}$ and replacing the gradient with the form in Eq.~\eqref{eq_ncsn_3} leads to training $s_\theta(x_t,\sigma_t)$ by minimizing the following objective, $\forall t \in \{1,\dots,T \}$:
 \begin{equation}
    \label{eq_ncsn_4}
    \begin{split}
    \mathcal{L}_{\scriptsize{\mbox{\emph{dsm}}}}\!=\! \frac{1}{T} \sum_{t=1}^{T} \lambda(\sigma_t) \mathbb{E}_{p(x)}\mathbb{E}_{x_t \sim p_{\sigma_t}(x_t|x)}\!\left\lVert s_\theta(x_t,\sigma_t)\!+\! \frac{x_t-x}{\sigma_t^2}\right\rVert_2^2\!,
    \end{split}
\end{equation}
where $\lambda(\sigma_t)$ is a weighting function. After training, the neural network $ s_\theta(x_t,\sigma_t)$ will return estimates of the scores $\nabla_{x_t} \log(p_{\sigma_t}(x_t))$, having as input the noisy image $x_t$ and the corresponding time step $t$.

At inference time, Song \etal~\cite{song-NeurIPS-2019} introduce the annealed Langevin dynamics, formally described in Algorithm \ref{annealed_langevin_dynamics}. Their method starts with white noise and applies Eq.~\eqref{eq_ncsn_1} for a fixed number of iterations. The required gradient (score) is given by the trained neural network conditioned on the time step $T$. The process continues for the following time steps, propagating the output of one step as input to the next. The final sample is the output returned for $t=0$.

\begin{algorithm}[t]
\caption{Annealed Langevin dynamics\label{annealed_langevin_dynamics}}
\textbf{Input}:

$\sigma_1, \dots, \sigma_T$ -- a sequence of Gaussian noise scales.

$N$ -- the number of Langevin dynamics iterations.

$\gamma_1, \dots, \gamma_T$ -- the update magnitudes for each noise scale.
\vspace{0.2em}
\algrule
\vspace{0.2em}
\textbf{Output}: 

$x_0^0$ -- the sampled image.
\vspace{0.2em}
\algrule
\vspace{0.2em}
\textbf{Computation}:
\begin{algorithmic}[1]
\State $x_T^0 \sim \mathcal{N}(0, \mathbf{I})$
\For {$t = T, \dots, 1$}
    \For {$i = 1, \dots, N$}
        \State $\omega \sim \mathcal{N}(0, \mathbf{I})$
        \State $x_t^i = x_t^{i-1} + \frac{\gamma_t}{2} \cdot s_\theta(x_t^{i-1},\sigma_t) + \sqrt{\gamma_t}\cdot\omega$
    \EndFor
    \State $x_{t-1}^0=x_t^N$
\EndFor
\end{algorithmic}
\end{algorithm}

\subsection{Stochastic Differential Equations (SDEs)}

Similar to the previous two methods, the approach presented in \cite{song-ICLR-2021} gradually transforms the data distribution $p(x_0)$ into noise. However, it generalizes over the previous two methods because, in its case, the diffusion process being considered to be continuous, thus becoming the solution of a stochastic differential equation (SDE). As shown in \cite{Andreson-SPA-1982}, the reverse process of this diffusion can be modeled with a reverse-time SDE which requires the score function of the density at each time step. Therefore, the generative model of Song \etal~\cite{song-ICLR-2021} employs a neural network to estimate the score functions, and generates samples from $p(x_0)$ by employing numerical SDE solvers. As in the case of NCSNs, the neural network receives the perturbed data and the time step as input, and produces an estimation of the score function.

The SDE of the forward diffusion process $(x_t)_{t=0}^{T}$, $t \in [0, T]$ has the following form:
\begin{equation}
    \begin{split}
    \label{forward_sde}
     \frac{\partial x}{\partial t}\!=\!f(x, t)\!+\! \sigma(t)\!\cdot\!\omega_t \Longleftrightarrow \partial x\!=\!f(x, t)\!\cdot\!\partial t\!+\! \sigma(t)\!\cdot\!\partial\omega,
    \end{split}
\end{equation}
where $\omega_t$ is Gaussian noise, $f$ is a function of $x$ and $t$ that computes the drift coefficient, and $\sigma$ is a time-dependent function that computes the diffusion coefficient. In order to have a diffusion process as a solution for this SDE, the drift coefficient should be designed such that it gradually nullifies the data $x_0$, while the diffusion coefficient controls how much Gaussian noise is added. 
The associated reverse-time SDE \cite{Andreson-SPA-1982} is defined as follows:
\begin{equation}
    \label{reverse_sde}
    \partial x = \left[f(x, t) - \sigma(t)^2\cdot \nabla_x \log{p_t(x)}\right]\cdot\partial t + \sigma(t)\cdot\partial\hat{\omega},
\end{equation}
where $\hat{\omega}$ represents the Brownian motion when the time is reversed, from $T$ to $0$. The reverse-time SDE shows that, if we start with pure noise, we can recover the data by removing the drift responsible for data destruction. The removal is performed by subtracting $ \sigma(t)^2\cdot \nabla_x \log{p_t(x)}$.

We can train the neural network $s_\theta(x,t)\approx\nabla_x \log{p_t(x)}$ by optimizing the same objective as in Eq.~\eqref{eq_ncsn_4}, but adapted for the continuous case, as follows:
\begin{equation}
    \label{eq_sde_dsm}
    \begin{split}
    &\mathcal{L}_{\scriptsize{\mbox{\emph{dsm}}}}^* = \\ 
    &=\!\mathbb{E}_t\!\left[\!\lambda(t)\mathbb{E}_{p(x_0)}\mathbb{E}_{p_t(x_t|x_0)}\!\left\lVert s_\theta(x_t,t)\!-\! \nabla_{x_t} \log{p_t(x_t|x_0)}\right\rVert_2^2\right]\!,
    \end{split}
\end{equation}
where $\lambda$ is a weighting function, and $t \sim \mathcal{U}([0,T])$. We underline that, when the drift coefficient $f$ is affine, $p_t(x_t|x_0)$ is a Gaussian distribution. When $f$ does not conform to this property, we cannot use denoising score matching, but we can fallback to sliced score matching \cite{Song-UAI-2019}.

The sampling for this approach can be performed with any numerical method applied on the SDE defined in Eq.~\eqref{reverse_sde}. In practice, the solvers do not work with the continuous formulation. For example, the Euler-Maruyama method fixes a tiny negative step $\Delta t$ and executes Algorithm \ref{alg_EuMa} until the initial time step $t=T$ becomes $t=0$. At step 3, the Brownian motion is given by $\Delta\hat{\omega} = \sqrt{|\Delta t|}\cdot z$, where $z \sim \mathcal{N}(0,\mathbf{I})$.

\begin{algorithm}[t]
\caption{Euler-Maruyama sampling method\label{alg_EuMa}}

\textbf{Input}: 

$\Delta t < 0$ -- a negative step close to $0$.

$f$ -- a function of $x$ and $t$ that computes the drift coefficient.

$\sigma$ -- a time-dependent function that computes the diffusion coefficient.

$\nabla_x \log{p_t(x)}$ -- the (approximated) score function.

$T$ -- the final time step of the forward SDE.
\vspace{0.2em}
\algrule
\vspace{0.2em}
\textbf{Output}: 

$x$ -- the sampled image.
\vspace{0.2em}
\algrule
\vspace{0.2em}
\textbf{Computation}:
\begin{algorithmic}[1]
\State $t = T$
\While{$t > 0$}
    \State $\Delta x\!=\!\left[f(x, t) - \sigma(t)^2\!\cdot\!\nabla_x \log{p_t(x)}\right]\!\cdot\!\Delta t + \sigma(t)\!\cdot\!\Delta\hat{\omega}$
   \State $x = x + \Delta x$
   \State $t = t + \Delta t$
\EndWhile
\end{algorithmic}
\end{algorithm}

% \begin{equation}
%     \label{eq_euler_maruyama}
%     \begin{split}
%     \Delta x &= \left[f(x, t) - \sigma(t)^2\cdot \nabla_x \log{p_t(x)}\right]\Delta t + \sigma(t)\Delta\hat{\omega}, \\
%     x &= x + \Delta x, \\
%     t &= t + \Delta t,
%     \end{split}
% \end{equation}

Song \etal~\cite{song-ICLR-2021} present several contributions in terms of sampling techniques. They introduce the Predictor-Corrector sampler which generates better examples. This algorithm first employs a numerical method to sample from the reverse-time SDE, and then uses a score-based method as a corrector, for example the annealed Langevin dynamics described in the previous subsection. Furthermore, they show that ordinary differential equations (ODEs) can also be used to model the reverse process. Hence, another sampling strategy unlocked by the SDE interpretation is based on numerical methods applied to ODEs. The main advantage of this latter strategy is its efficiency.

\subsection{Relation to Other Generative Models}

We discuss below the connections between diffusion models and other types of generative models. We start with likelihood-based methods and finish with generative adversarial networks.

Diffusion models have more aspects in common with VAEs \cite{Kingma-ICLR-2014b}. For instance, in both cases, the data is mapped to a latent space and the generative process learns to transform the latent representations into data. Moreover, in both situations, the objective function can be derived as a lower-bound of the data likelihood. Nevertheless, there are essential differences between the two approaches and, further, we will mention some of them. The latent representation of a VAE contains compressed information about the original image, while diffusion models destroy the data entirely after the last step of the forward process. The latent representations of diffusion models have the same dimensions as the original data, while VAEs work better when the dimensions are reduced. Ultimately, the mapping to the latent space of a VAE is trainable, which is not true for the forward process of diffusion models because, as stated before, the latent is obtained by gradually adding Gaussian noise to the original image. The aforementioned similarities and differences can be the key for future developments of the two methods. For example, there already exists some work that builds more efficient diffusion models by applying them on the latent space of a VAE \cite{pandey-NeurIPSW-2021, vahdat-NeurIPS-2021}.

Autoregressive models \cite{Oord-NeurIPS-2016, salimans-ICLR-2017} represent images as sequences of pixels. Their generative process produces new samples by generating an image pixel by pixel, conditioned on the previously generated pixels. This approach implies a unidirectional bias that clearly represents a limitation of this class of generative models. Esser \etal~\cite{esser-NeurIPS-2021} see diffusion and autoregressive models as complementary and solve the above issue. Their method learns to reverse a multinomial diffusion process via a Markov chain where each transition is implemented as an autoregressive model. The global information provided to the autoregressive model is given by the previous step of the Markov chain.

Normalizing flows \cite{dinh-ICLR-2015, dinh-ICLR-2017} are a class of generative models that transform a simple Gaussian distribution into a complex data distribution. The transformation is done via a set of invertible functions which have an easy-to-compute Jacobian determinant. These conditions translate in practice into architectural restrictions. An important feature of this type of model is that the likelihood is tractable. Hence, the objective for training is the negative log-likelihood.
When comparing with diffusion models, the two types of models have in common the mapping of the data distribution to Gaussian noise. However, the similarities between the two methods end here, because normalizing flows perform the mapping in a deterministic fashion by learning an invertible and differentiable function. These properties imply, in contrast to diffusion models, additional constraints on the network architecture, and a learnable forward process. A method which connects these two generative algorithms is DiffFlow. Introduced in \cite{zhang-NeurIPS-2021}, DiffFlow extends both diffusion models and normalizing flows such that the reverse and forward processes are both trainable and stochastic.

Energy-based models (EBMs) \cite{lecun-PSD-2006, ngiam-ICML-2011, swersky-ICML-2011, bao-ICML-2021} focus on providing estimates of unnormalized versions of density functions, called energy functions. Thanks to this property and in contrast to the previous likelihood-based methods, this type of model can be represented with any regression neural network. However, due to this flexibility, the training  of EBMs is difficult. One popular training strategy used in practice is score matching \cite{swersky-ICML-2011, bao-ICML-2021}. Regarding the sampling, among other strategies, there is the Markov Chain Monte Carlo (MCMC) method, which is based on the score function. Therefore, the formulation from Subsection~\ref{sec_NCSN} of diffusion models can be considered to be a particular case of the energy-based framework, precisely the case when the training and sampling only require the score function.

GANs \cite{Goodfellow-NIPS-2014} were considered by many as state-of-the-art generative models in terms of the quality of the generated samples, before the recent rise of diffusion models \cite{dhariwal-NeurIPS-2021}. GANs are also known as being difficult to train due to their adversarial objective \cite{salimans-NeurIPS-16}, and often suffer from mode collapse. In contrast, diffusion models have a stable training process and provide more diversity because they are likelihood-based. Despite these advantages, diffusion models are still inefficient when compared to GANs, requiring multiple network evaluations during inference. A key aspect for comparison between GANs and diffusion models is their latent space. While GANs have a low-dimensional latent space, diffusion models preserve the original size of the images. Furthermore, the latent space of diffusion models is usually modeled as a random Gaussian distribution, being similar to VAEs. In terms of semantic properties, it was discovered that the latent space of GANs contains subspaces associated with visual attributes \cite{Shen-CVPR-2020}. Thanks to this property, the attributes can be manipulated with changes in the latent space \cite{Radford-ICLR-2016, Shen-CVPR-2020}. In contrast, when such transformations are desired for diffusion models, the preferred procedure is the guidance technique \cite{ho-NeurIPS-2021, dhariwal-NeurIPS-2021}, which does not exploit any semantic property of the latent space. However, Song \etal~\cite{song-ICLR-2021} demonstrate that the latent space of diffusion models has a well-defined structure, illustrating that interpolations in this space lead to interpolations in the image space. In summary, from the semantic perspective, the latent space of diffusion models has been explored much less than in the case of GANs, but this may be one of the future research directions to be followed by the community.

\section{A Categorization of Diffusion Models}

We categorize diffusion models into a multi-perspective taxonomy considering different criteria of separation. Perhaps the most important criteria to separate the models are defined by $(i)$ the task they are applied to, and $(ii)$ the input signals they require. Furthermore, as there are multiple approaches in formulating a diffusion model, $(iii)$ the underlying framework is another key factor for classifying diffusion models. Finally, the $(iv)$ data sets used during training and evaluation are also of high importance, because they provide the means to compare different models on the same task. Our categorization of diffusion models according to the criteria enumerated above is presented in Table \ref{tab_taxonomy}. 

In the remainder of this section, we present several contributions on diffusion models, choosing the target task as the primary criterion to separate the methods. We opted for this classification criterion as it is fairly well-balanced and representative for research on diffusion models, facilitating a quick grasping of related works by readers working on specific tasks. Although the main task is usually related to image generation, a considerable amount of work has been conducted to match and even surpass the performance of GANs on other topics, such as super-resolution, inpainting, image editing, image-to-image translation or segmentation. 

\begin{table*}[!th]
\setlength\tabcolsep{2pt}
\caption{Our multi-perspective categorization of diffusion models applied in computer vision. To classify existing models, we consider three criteria: the task, the denoising condition, and the underlying approach (architecture). Additionally, we list the data sets on which the surveyed models are applied.
We use the following abbreviations in the architecture column: D3PM (Discrete Denoising Diffusion Probabilistic Models), DSB (Diffusion Schr{\"o}dinger Bridge), BDDM (Bilateral Denoising Diffusion Models), PNDM (Pseudo Numerical Methods for Diffusion Models), ADM (Ablated Diffusion Model), D2C (Diffusion-Decoding Models with Contrastive Representations), CCDF (Come-Closer-Diffuse-Faster), VQ-DDM (Vector Quantised Discrete Diffusion Model), BF-CNN (Bias-Free CNN), FDM (Flexible Diffusion Model), RVD (Residual Video Diffusion), RaMViD (Random Mask Video Diffusion).} 
\label{tab_taxonomy}
\centering
\begin{tabular}{|p{3.2cm}|p{4.4cm}|p{3.5cm}|p{2.2cm}|p{4.0cm}|}
\hline
\textbf{Paper}  &   \textbf{Task}  &   \textbf{Denoising Condition} &   \textbf{Architecture}  &   \textbf{Data Sets}  \\ 
\hline
%15
\rowcolor{LightRed}
Austin \etal~\cite{austin-NeurIPS-2021} & image generation & unconditional & D3PM & CIFAR-10 \\
\hline
%56
\rowcolor{LightRed}
Bao \etal~\cite{bao-arXiv-2022} & image generation & unconditional & DDIM, Improved DDPM & CelebA, ImageNet, LSUN Bedroom, CIFAR-10 \\
\hline
%42
\rowcolor{LightRed}
Benny \etal~\cite{benny-CVPR-2022} &  image generation & unconditional & DDPM, DDIM & CIFAR-10, ImageNet, CelebA \\ 
\hline
%29
\rowcolor{LightRed}
Bond-Taylor \etal~\cite{bond-ECCV-2022} &  image generation & unconditional & DDPM & LSUN Bedroom, LSUN Church, FFHQ \\ 
\hline
%48
\rowcolor{LightRed}
Choi \etal~\cite{choi-CVPR-2022} &  image generation & unconditional & DDPM & FFHQ, AFHQ-Dog, CUB, MetFaces \\ 
\hline
%14
\rowcolor{LightRed}
De \etal~\cite{de-NeurIPS-2021} &  image generation & unconditional & DSB & MNIST, CelebA  \\
\hline
%67
\rowcolor{LightRed}
Deasy \etal~\cite{deasy-arXiv-2021} &  image generation & unconditional & NCSN & MNIST, Fashion-MNIST, CIFAR-10, CelebA \\ 
\hline
%82
\rowcolor{LightRed}
Deja \etal~\cite{deja-arXiv-2022} &  image generation & unconditional & Improved DDPM & Fashion-MNIST, CIFAR-10, CelebA  \\ 
\hline
%58
\rowcolor{LightRed}
Dockhorn \etal~\cite{dockhorn-arXiv-2021} &  image generation & unconditional & NCSN++, DDPM++ & CIFAR-10 \\ 
\hline
%3
\rowcolor{LightRed}
Ho \etal~\cite{ho-NeurIPS-2020} &  image generation & unconditional & DDPM &  CIFAR-10, CelebA-HQ, LSUN \\ 
\hline
%16
\rowcolor{LightRed}
Huang \etal~\cite{huang-NeurIPS-2021} &  image generation & unconditional & DDPM & CIFAR-10, MNIST \\
\hline
%76
\rowcolor{LightRed}
Jing \etal~\cite{jing-arXiv-2022} &  image generation & unconditional & NCSN++, DDPM++ & CIFAR-10, CelebA-256-HQ, LSUN Church \\ 
\hline
%5
\rowcolor{LightRed}
Jolicoeur \etal~\cite{jolicoeur-ICLR-2021} &  image generation & unconditional & NCSN & CIFAR-10, LSUN Church, Stacked-MNIST \\ 
\hline
%33
\rowcolor{LightRed}
Jolicoeur \etal~\cite{jolicoeur-arXiv-2021} &  image generation & unconditional & DDPM++, NCSN++ & CIFAR-10, LSUN Church, FFHQ \\ 
\hline
%79
\rowcolor{LightRed}
Kim \etal~\cite{kim-arXiv-2022} &  image generation & unconditional & NCSN++, DDPM++ & CIFAR-10, CelebA, MNIST \\ 
\hline
%55
\rowcolor{LightRed}
Kingma \etal~\cite{kingma-NeurIPS-2021} &  image generation & unconditional & DDPM & CIFAR-10, ImageNet \\ 
\hline
%25
\rowcolor{LightRed}
Kong \etal~\cite{kong-ICMLW-2021} &  image generation & unconditional & DDIM, DDPM & LSUN Bedroom, CelebA, CIFAR-10 \\ 
\hline
%27
\rowcolor{LightRed}
Lam \etal~\cite{lam-arXiv-2021} &  image generation & unconditional & BDDM  & CIFAR-10, CelebA \\ 
\hline
%39
\rowcolor{LightRed}
Liu \etal~\cite{liu-ICLR-2022} &  image generation & unconditional & PNDM  & CIFAR-10, CelebA \\
\hline
%86
\rowcolor{LightRed}
Ma \etal~\cite{ma-arXiv-2022} &  image generation & unconditional & NCSN, NCSN++ & CIFAR-10, CelebA, LSUN Bedroom, LSUN Church, FFHQ  \\ 
%26
\hline
\rowcolor{LightRed}
Nachmani \etal~\cite{nachmani-arXiv-2021} &  image generation & unconditional & DDIM, DDPM & CelebA, LSUN Church \\ 
\hline
%21
\rowcolor{LightRed}
Nichol \etal~\cite{nichol-ICML-2021} &  image generation & unconditional & DDPM & CIFAR-10, ImageNet \\ 
\hline
%37
%38
\rowcolor{LightRed}
Pandey \etal~\cite{pandey-NeurIPSW-2021} &  image generation & unconditional & DDPM & CelebA-HQ, CIFAR-10 \\ 
\hline
%32
\rowcolor{LightRed}
San \etal~\cite{san-arXiv-2021} &  image generation & unconditional & DDPM & CelebA, LSUN Bedroom, LSUN Church \\ 
\hline
%45
\rowcolor{LightRed}
Sehwag \etal~\cite{sehwag-CVPR-2022} &  image generation & unconditional & ADM & CIFAR-10, ImageNet\\ 
\hline
%1
\rowcolor{LightRed}
Sohl-Dickstein \etal~\cite{sohl-icml-2015} &  image generation & unconditional & DDPM & MNIST, CIFAR-10, Dead Leaf Images \\ 
\hline
%4
\rowcolor{LightRed}
Song \etal~\cite{song-NeurIPS-2020} &  image generation & unconditional & NCSN & FFHQ, CelebA, LSUN Bedroom, LSUN Tower, LSUN Church Outdoor \\ 
\hline
%9
\rowcolor{LightRed}
Song \etal~\cite{song-NeurIPS-2021} &  image generation & unconditional & DDPM++ & CIFAR-10, ImageNet 32$\times$32 \\ 
\hline
%22
\rowcolor{LightRed}
Song \etal~\cite{song-ICLR-2021b} &  image generation & unconditional & DDIM & CIFAR-10, CelebA, LSUN \\ 
\hline
%19
\rowcolor{LightRed}
Vahdat \etal~\cite{vahdat-NeurIPS-2021} &  image generation & unconditional & NCSN++ & CIFAR-10, CelebA-HQ, MNIST \\ 
\hline
%35
\rowcolor{LightRed}
Wang \etal~\cite{wang-ICML-2021} &  image generation & unconditional & DDIM & CIFAR-10, CelebA \\ 
\hline
%85
\rowcolor{LightRed}
Wang \etal~\cite{wang-arXiv-2022a} &  image generation & unconditional & StyleGAN2, ProjectedGAN & CIFAR-10, STL-10, LSUN Bedroom, LSUN Church, AFHQ, FFHQ \\ 
\hline
%34
\rowcolor{LightRed}
Watson \etal~\cite{watson-arXiv-2021} &  image generation & unconditional & DDPM & CIFAR-10, ImageNet \\ 
\hline
%69
\rowcolor{LightRed}
Watson \etal~\cite{watson-ICLR-2021} &  image generation & unconditional & Improved DDPM & CIFAR-10, ImageNet 64$\times$64 \\ 
\hline
%59
\rowcolor{LightRed}
Xiao \etal~\cite{xiao-arXiv-2021} &  image generation & unconditional & NCSN++ & CIFAR-10 \\ 
\hline
%13
\rowcolor{LightRed}
Zhang \etal~\cite{zhang-NeurIPS-2021} &  image generation & unconditional & DDPM & CIFAR-10, MNIST \\ 
\hline
%71
\rowcolor{LightRed}
Zheng \etal~\cite{zheng-arXiv-2022} &  image generation & unconditional & DDPM & CIFAR-10, CelebA, CelebA-HQ, LSUN Bedroom, LSUN Church \\ 
\hline
%66
\rowcolor{LightOrange}
Bordes \etal~\cite{bordes-arXiv-2021} & conditional image generation & conditioned on latent representations & Improved DDPM & ImageNet \\ 
\hline
%81
\rowcolor{LightOrange}
Campbell \etal~\cite{campbell-arXiv-2022} & conditional image generation & unconditional, conditioned on sound & DDPM & CIFAR-10, Lakh Pianoroll \\ 
\hline
%73
\rowcolor{LightOrange}
Chao \etal~\cite{chao-arXiv-2022} & conditional image generation & conditioned on class & Score SDE, Improved DDPM & CIFAR-10, CIFAR-100 \\ 
\hline
%11
\rowcolor{LightOrange}
Dhariwal \etal~\cite{dhariwal-NeurIPS-2021} & conditional image generation & unconditional, classifier guidance & ADM & LSUN Bedroom, LSUN Horse, LSUN Cat \\ 
\hline
\end{tabular}
\end{table*}

\begin{table*}[!th]
\setlength\tabcolsep{2pt}
\centering
\begin{tabular}{|p{3.2cm}|p{4.4cm}|p{3.5cm}|p{2.2cm}|p{4.0cm}|}
\hline
%50
%53
\rowcolor{LightOrange}
Ho \etal~\cite{ho-arXiv-2021} & conditional image generation & conditioned on label & DDPM & LSUN, ImageNet\\ 
\hline
%24
\rowcolor{LightOrange}
Ho \etal~\cite{ho-NeurIPS-2021} & conditional image generation & unconditional, classifier-free guidance & ADM & ImageNet 64$\times$64, ImageNet 128$\times$128 \\
\hline
%57
\rowcolor{LightOrange}
Karras \etal~\cite{karras-arXiv-2022} & conditional image generation & unconditional, conditioned on class & DDPM++, NCSN++, DDPM, DDIM & CIFAR-10, ImageNet 64$\times$64 \\ 
\hline
%31
\rowcolor{LightOrange}
Liu \etal~\cite{liu-arXiv-2021} & conditional image generation & conditioned on text, image, style guidance & DDPM & FFHQ, LSUN Cat, LSUN Horse, LSUN Bedroom \\ 
\hline
%106
\rowcolor{LightOrange}
Liu \etal~\cite{liu-arXiv-2022b} & conditional image generation & conditioned on text, 2D positions, relational descriptions between items, human facial attributes & Improved DDPM & CLEVR, Relational CLEVR, FFHQ  \\ 
\hline
%83
%84
\rowcolor{LightOrange}
Lu \etal~\cite{lu-arXiv-2022} & conditional image generation & unconditional, conditioned on class & DDIM & CIFAR-10, CelebA, ImageNet, LSUN Bedroom \\
\hline
%60
\rowcolor{LightOrange}
Salimans \etal~\cite{salimans-arXiv-2022} & conditional image generation & unconditional, conditioned on class & DDIM & CIFAR-10, ImageNet, LSUN \\ 
\hline
%77
\rowcolor{LightOrange}
Singh \etal~\cite{singh-arXiv-2022} & conditional image generation & conditioned on noise & DDIM & ImageNet \\
\hline
%12
\rowcolor{LightOrange}
Sinha \etal~\cite{sinha-NeurIPS-2021} & conditional image generation & unconditional, conditioned on label & D2C & CIFAR-10, CIFAR-100, fMoW, CelebA-64, CelebA-HQ-256, FFHQ-256 \\ 
\hline
%61
\rowcolor{LightYellow}
Ho \etal~\cite{saharia-SIGGRAPH-2022} & image-to-image translation & conditioned on image & Improved DDPM & ctest10k, places10k \\
\hline
%96
\rowcolor{LightYellow}
Li \etal~\cite{li-arXiv-2022} & image-to-image translation & conditioned on image & DDPM & Face2Comic, Edges2Shoes, Edges2Handbags \\ 
\hline
%101
\rowcolor{LightYellow}
Sasaki \etal~\cite{sasaki-arXiv-2021} & image-to-image translation & conditioned on image & DDPM & CMP Facades, KAIST Multispectral Pedestrian \\
\hline
%95
\rowcolor{LightYellow}
Wang \etal~\cite{wang-arXiv-2022c} & image-to-image translation & conditioned on image & DDIM & ADE20K, COCO-Stuff, DIODE  \\ 
\hline
%98
\rowcolor{LightYellow}
Wolleb \etal~\cite{wolleb-arXiv-2022b} & image-to-image translation & conditioned on image & Improved DDPM & BRATS \\ 
\hline
%94
\rowcolor{LightYellow}
Zhao \etal~\cite{zhao-arXiv-2022} & image-to-image translation & conditioned on image & DDPM & CelebaA-HQ, AFHQ  \\ 
\hline
%107
\rowcolor{LightGreen}
Gu \etal~\cite{gu-CVPR-2022} & text-to-image generation & conditioned on text & VQ-Diffusion & CUB-200, Oxford 102 Flowers, MS-COCO \\ 
\hline
\rowcolor{LightGreen}
Jiang \etal~\cite{jiang-arXiv-2022} & text-to-image generation & conditioned on text & Transformer-based encoder-decoder & DeepFashion-MultiModal \\ 
\hline
%52
\rowcolor{LightGreen}
Ramesh \etal~\cite{ramesh-arXiv-2022} & text-to-image generation & conditioned on text & ADM & MS-COCO, AVA \\ 
\hline
%105
\rowcolor{LightGreen}
Rombach \etal~\cite{rombach-arXiv-2022} & text-to-image generation & conditioned on text & LDM & OpenImages, WikiArt, LAION-2B-en, ArtBench \\ 
\hline
%54
\rowcolor{LightGreen}
Saharia \etal~\cite{saharia-arXiv-2022} & text-to-image generation & conditioned on text  & Imagen & MS-COCO, DrawBench \\
\hline
%104
\rowcolor{LightGreen}
Shi \etal~\cite{shi-arXiv-2022b} & text-to-image generation & unconditional, conditioned on text & Improved DDPM & Conceptual Captions, MS-COCO \\ 
\hline
%75
\rowcolor{LightGreen}
Zhang \etal~\cite{zhang-arXiv-2022} & text-to-image generation & unconditional, conditioned on text & DDIM & CIFAR-10, CelebA, ImageNet \\ 
\hline
%36
\rowcolor{LightCyan}
Daniels \etal~\cite{daniels-NeurIPS-2021} & super-resolution & conditioned on image & NCSN & CIFAR-10, CelebA \\ 
\hline
%20
\rowcolor{LightCyan}
Saharia \etal~\cite{saharia-arXiv-2021} & super-resolution & conditioned on image & DDPM++ & FFHQ, CelebA-HQ, ImageNet-1K \\ 
\hline
%102
%103
\rowcolor{LightBlue}
Avrahami \etal~\cite{avrahami-arXiv-2022} & image editing & conditioned on image and mask & DDPM, ADM & ImageNet, CUB, LSUN Bedroom, MS-COCO \\ 
\hline
%51
\rowcolor{LightBlue}
Avrahami \etal~\cite{avrahami-CVPR-2022} & region image editing & text guidance & DDPM & PaintByWord \\ 
\hline
%63
\rowcolor{LightBlue}
Meng \etal~\cite{meng-arXiv-2021} & image editing & conditioned on image & Score SDE, DDPM, Improved DDPM & LSUN, CelebA-HQ \\ 
\hline
%41
\rowcolor{LightIndigo}
Lugmayr \etal~\cite{lugmayr-CVPR-2022} & inpainting & unconditional & DDPM & CelebA-HQ, ImageNet \\ 
\hline
%17
\rowcolor{LightIndigo}
Nichol \etal~\cite{nichol-arXiv-2021} & inpainting & conditioned on image, text guidance & ADM & MS-COCO \\ 
\hline
%90
\rowcolor{LightMagenta}
Amit \etal~\cite{amit-arXiv-2021} & image segmentation & conditioned on image & Improved DDPM & Cityscapes, Vaihingen, MoNuSeg \\ 
\hline
%64
\rowcolor{LightMagenta}
Baranchuk \etal~\cite{baranchuk-arXiv-2021} & image segmentation & conditioned on image & Improved DDPM & LSUN, FFHQ-256, ADE-Bedroom-30, CelebA-19 \\ 
\hline
%30
\rowcolor{DirtyWhite}
Batzolis \etal~\cite{batzolis-arXiv-2021} & multi-task (inpainting, super-resolution, edge-to-image) & conditioned on image & DDPM & CelebA, Edges2Shoes \\ 
\hline
%92
\rowcolor{DirtyWhite}
Batzolis \etal~\cite{batzolis-arXiv-2022} & multi-task (image generation, super-resolution, inpainting, image-to-image translation) & unconditional & DDIM & ImageNet, CelebA-HQ, CelebA, Edges2Shoes \\ 
\hline
%74
\rowcolor{DirtyWhite}
Blattmann \etal~\cite{blattmann-arXiv-2022} & multi-task (image generation) & unconditional, conditioned on text, class & LDM & ImageNet \\
\hline
%65
\rowcolor{DirtyWhite}
Choi \etal~\cite{choi-arXiv-2021} & multi-task (image generation, image-to-image translation, image editing) & conditioned on image & DDPM & FFHQ, MetFaces \\ 
\hline
%47
\rowcolor{DirtyWhite}
Chung \etal~\cite{chung-CVPR-2022} & multi-task (inpainting, super-resolution, MRI reconstruction) & conditioned on image & CCDF & FFHQ, AFHQ, fastMRI knee \\ 
\hline
%10
\rowcolor{DirtyWhite}
Esser \etal~\cite{esser-NeurIPS-2021} & multi-task (image generation, inpainting) & unconditional, conditioned on class, image and text & ImageBART & ImageNet, Conceptual Captions, FFHQ, LSUN \\ 
\hline
%23
\rowcolor{DirtyWhite}
Gao \etal~\cite{gao-ICLR-2021} & multi-task (image generation, inpainting) & unconditional, conditioned on image & DDPM &  CIFAR-10, LSUN, CelebA \\
\hline 
%88
\rowcolor{DirtyWhite}
Graikos \etal~\cite{graikos-arXiv-2022} & multi-task (image generation, image segmentation) & conditioned on class & DDIM & FFHQ-256, CelebA \\ 
\hline
\end{tabular}
\end{table*}

\begin{table*}[!th]
\setlength\tabcolsep{2pt}
\centering
\begin{tabular}{|p{3.2cm}|p{4.4cm}|p{3.5cm}|p{2.2cm}|p{4.0cm}|}
\hline
%43
\rowcolor{DirtyWhite}
Hu \etal~\cite{hu-CVPR-2022} & multi-task (image generation, inpainting) & unconditional, conditioned on image & VQ-DDM & CelebA-HQ, LSUN Church \\ 
\hline
%70
\rowcolor{DirtyWhite}
Khrulkov \etal~\cite{khrulkov-arXiv-2022} & multi-task (image generation, image-to-image translation) & conditioned on class & Improved DDPM & AFHQ, FFHQ, MetFaces, ImageNet \\ 
\hline
%49
\rowcolor{DirtyWhite}
Kim \etal~\cite{kim-CVPR-2022} & multi-task (image translation, multi-attribute transfer) & conditioned on image, portrait, stroke & DDIM & ImageNet, CelebA-HQ, AFHQ-Dog, LSUN Bedroom, Church \\ 
\hline
%8
\rowcolor{DirtyWhite}
Luo \etal~\cite{luo-CVPR-2021} & multi-task (point cloud generation, auto-encoding, unsupervised representation learning) & conditioned on shape latent & DDPM & ShapeNet  \\ 
\hline
%78
\rowcolor{DirtyWhite}
Lyu \etal~\cite{lyu-arXiv-2022} & multi-task (image generation, image editing) & unconditional, conditioned on class & DDPM & CIFAR-10, CelebA, ImageNet, LSUN Bedroom, LSUN Cat \\ 
\hline
%46
\rowcolor{DirtyWhite}
Preechakul \etal~\cite{preechakul-CVPR-2022} & multi-task (latent interpolation, attribute manipulation) & conditioned on latent representation& DDIM  & CelebA-HQ \\ 
\hline
%44
\rowcolor{DirtyWhite}
Rombach \etal~\cite{rombach-CVPR-2022} & multi-task (super-resolution, image generation, inpainting) & unconditional, conditioned on image & VQ-DDM & ImageNet, CelebA-HQ, FFHQ, LSUN \\ 
\hline
%72
\rowcolor{DirtyWhite}
Shi \etal~\cite{shi-arXiv-2022a} & multi-task (super-resolution, inpainting) & conditioned on image & Improved DDPM & MNIST, CelebA \\ 
\hline
%2
\rowcolor{DirtyWhite}
Song \etal~\cite{song-NeurIPS-2019} & multi-task (image generation, inpainting) & unconditional, conditioned on image & NCSN & MNIST, CIFAR-10, CelebA \\ 
\hline
%112
\rowcolor{DirtyWhite}
Kadkhodaie \etal~\cite{kadkhodaie-NeurIPS-2021} & multi-task (Spatial super-resolution, Deblurring, Compressive sensing, 
Inpainting, Random missing pixels) & conditioned on linear measurements & BF-CNN & MNIST, Set5, Set68, Set14 \\
\hline
%6
%7
\rowcolor{DirtyWhite}
Song \etal~\cite{song-ICLR-2021} & multi-task (image generation, inpainting, colorization) & unconditional, conditioned on image, class & NCSN++, DDPM++ & CelebA-HQ, CIFAR-10, LSUN \\ 
\hline
%110
\rowcolor{LightRed}
Hu \etal~\cite{hu-SPIE-2022} & medical image-to-image translation & conditioned on image & DDPM & ONH \\ 
\hline
%109
\rowcolor{LightRed}
Chung \etal~\cite{chung-MIA-2022} & medical image generation & conditioned on measurements & NCSN++ & fastMRI knee \\ 
\hline
%93
\rowcolor{LightRed}
{\"O}zbey \etal~\cite{ozbey-arXiv-2022} & medical image generation & conditioned on image & Improved DDPM & IXI, Gold Atlas - Male Pelvis \\
\hline
%108
\rowcolor{LightRed}
Song \etal~\cite{song-arXiv-2021} & medical image generation & conditioned on measurements & NCSN++ & LIDC, LDCT Image and Projection, BRATS \\ 
\hline
%89
\rowcolor{LightRed}
Wolleb \etal~\cite{wolleb-arXiv-2021} & medical image segmentation & conditioned on image & Improved DDPM & BRATS \\ 
\hline
% Paper from Pedro (Uni of Edin)
\rowcolor{LightOrange}
Sanchez \etal~\cite{Sanchez-DGM4Miccai-2022} & medical image segmentation and anomaly detection & conditioned on image and binary variable & ADM & BRATS \\
\hline
%111
\rowcolor{LightOrange}
Pinaya \etal~\cite{pinaya-arXiv-2022} & medical image segmentation and anomaly detection & conditioned on image & DDPM & MedNIST, UK Biobank Images, WMH, BRATS, MSLUB \\ 
\hline
%97
\rowcolor{LightOrange}
Wolleb \etal~\cite{wolleb-arXiv-2022a} & medical image anomaly detection & conditioned on image & DDIM & CheXpert, BRATS \\ 
\hline
\rowcolor{LightOrange}
Wyatt \etal~\cite{Wyatt-CVPRW-2022} & medical image anomaly detection & conditioned on image & ADM & NFBS, 22 MRI scans\\
\hline
%video
\rowcolor{LightYellow}
Harvey \etal~\cite{harvey-arXiv-2022} & video generation & conditioned on frames & FDM & GQN-Mazes, MineRL Navigate, CARLA Town01 \\
\hline
%video
\rowcolor{LightYellow}
Ho \etal~\cite{ho-ICLRW-2022} & video generation & unconditional, conditioned on text & DDPM & 101 Human Actions\\
\hline
%video
\rowcolor{LightYellow}
Yang \etal~\cite{yang-arXiv-2022} & video generation & conditioned on video representation &  RVD & BAIR, KTH
Actions, Simulation, Cityscapes \\
\hline
%video
\rowcolor{LightYellow}
H{\"o}ppe \etal~\cite{hoppe-arXiv-2022} & video generation and infilling & conditioned on frames & RaMViD & BAIR, Kinetics-600, UCF-101\\
\hline
%80
\rowcolor{LightGreen}
Giannone \etal~\cite{giannone-arXiv-2022} & few-shot image generation & conditioned on image & Improved DDPM & CIFAR-FS, mini-ImageNet, CelebA \\ 
\hline
%40
\rowcolor{LightCyan}
Jeanneret \etal~\cite{jeanneret-arXiv-2022} & counterfactual explanations & unconditional & DDPM & CelebA \\ 
\hline
\rowcolor{LightCyan}
Sanchez \etal~\cite{sanchez-CLeaR-2022} & counterfactual estimates & conditional & ADM & MNIST, ImageNet \\
\hline
%99
\rowcolor{LightBlue}
Kawar \etal~\cite{kawar-arXiv-2022} & image restoration & conditioned on image & DDIM & FFHQ, ImageNet \\ 
\hline
%91
\rowcolor{LightBlue}
{\"O}zdenizci \etal~\cite{ozdenizci-arXiv-2022} & image restoration & conditioned on image & DDPM & Snow100K, Outdoor-Rain, RainDrop \\ 
\hline
%100
\rowcolor{LightIndigo}
Kim \etal~\cite{kim-arXiv-2021} & image registration & conditioned on image & DDPM & Radboud Faces, OASIS-3 \\ 
\hline
%62
\rowcolor{LightMagenta}
Nie \etal~\cite{nie-arXiv-2022} & adversarial purification & conditioned on image & Score SDE, Improved DDPM, DDIM & CIFAR-10, ImageNet, CelebA-HQ \\ 
\hline
%87
\rowcolor{LightRed}
Wang \etal~\cite{wang-arXiv-2022b} & semantic image generation & conditioned on semantic map & DDPM & Cityscapes, ADE20K, CelebAMask-HQ \\ 
\hline
%18
\rowcolor{LightOrange}
Zhou \etal~\cite{zhou-ICCV-2021} & shape generation and completion & unconditional, conditional shape completion & DDPM & ShapeNet, PartNet \\ 
\hline
%28
\rowcolor{LightYellow}
Zimmermann \etal~\cite{zimmermann-arXiv-2021} & classification & conditioned on label & DDPM++ & CIFAR-10 \\
\hline
\end{tabular}
\end{table*}

\subsection{Unconditional Image Generation}

The diffusion models presented below are used to generate samples in an unconditional setting. Such models do not require supervision signals, being completely unsupervised. We consider this as the most basic and generic setting for image generation.

\subsubsection{Denoising Diffusion Probabilistic Models}

% 1 Deep Unsupervised Learning using Nonequilibrium Thermodynamics
The work of Sohl-Dickstein \etal~\cite{sohl-icml-2015} formalizes diffusion models as described in Section \ref{sec2.1}. The proposed neural network is based on a convolutional architecture containing multi-scale convolution. %and their work presents the results of image generation experiments on the following data sets: MNIST, CIFAR-10 and Dead Leaf Images.

%15. Structured Denoising Diffusion Models in Discrete State-Spaces
Austin \etal~\cite{austin-NeurIPS-2021} extend the approach of Sohl-Dickstein \etal~\cite{sohl-icml-2015} to discrete diffusion models, studying different choices for the transition matrices used in the forward process. Their results are competitive with previous continuous diffusion models for the image generation task. % applied on CIFAR-10.

% 3.Denoising Diffusion Probabilistic Models 
Ho \etal~\cite{ho-NeurIPS-2020} extend the work presented in \cite{sohl-icml-2015}, proposing to learn the reverse process by estimating the noise in the image at each step. This change leads to an objective that resembles the denoising score matching applied in \cite{song-NeurIPS-2019}. %The authors test their method on image generation using CIFAR-10, CelebA-HQ, LSUN Church and Bedroom data sets. 
To predict the noise in an image, the authors use the PixelCNN++ architecture, which was introduced in \cite{salimans-ICLR-2017}.

% 21.Improved Denoising Diffusion Probabilistic Models
On top of the work proposed by Ho \etal~\cite{ho-NeurIPS-2020}, Nichol \etal \cite{nichol-ICML-2021} introduce several improvements, observing that the linear noise schedule is suboptimal for low resolution. They propose a new option that avoids a fast information destruction towards the end of the forward process. Further, they show that it is required to learn the variance in order to improve the performance of diffusion models in terms of log-likelihood. This last change allows faster sampling, somewhere around 50 steps being required. %The image generation experiments were conducted on CIFAR-10 and ImageNet and the architecture was similar to DDPM \cite{ho-NeurIPS-2020}.

% 22. Denoising Diffusion Implicit Models
Song \etal~\cite{song-ICLR-2021b} replace the Markov forward process used in \cite{ho-NeurIPS-2020} with a non-Markovian one. The generative process changes such that the model first predicts the normal sample, and then, it is used to estimate the next step in the chain. The change leads to a faster sampling procedure with a small impact on the quality of the generated samples. The resulting framework is known as the denoising diffusion implicit model (DDIM).% neural network architecture is the same as in \cite{ho-NeurIPS-2020} and it was used for image generation experiments on CIFAR-10, CelebA and LSUN data sets.

%12. D2C: Diffusion-Decoding Models for Few-Shot Conditional Generation
The work of Sinha \etal~\cite{sinha-NeurIPS-2021} presents the diffusion-decoding model with contrastive representations (D2C), a generative method which trains a diffusion model on latent representations produced by an encoder. The framework, which is based on the DDPM architecture presented in \cite{ho-NeurIPS-2020}, produces images by mapping the latent representations to images. 

% 32. Noise Estimation for Generative Diffusion Models
In \cite{san-arXiv-2021}, the authors present a method to estimate the noise parameters given the current input at inference time. Their change improves the Fr\'{e}chet Inception Distance (FID), while requiring less steps. The authors employ VGG-11 to estimate the noise parameters, and DDPM \cite{ho-NeurIPS-2020} to generate images. %They conduct the experiments on the following data sets: CelebA, LSUN Bedroom and Church.

% 26. Non Gaussian Denoising Diffusion Models
The work of Nachmani \etal~\cite{nachmani-arXiv-2021} suggests replacing the Gaussian noise distributions of the diffusion process with two other distributions, a mixture of two Gaussians and the Gamma distribution. The results show better FID values and faster convergence thanks to the Gamma distribution that has higher modeling capacity. %The experiments are conducted on CelebA and LSUN Church data sets using the previous models introduced in \cite{song-ICLR-2021b, ho-NeurIPS-2020}.

% 27. Bilateral Denoising Diffusion Models
Lam \etal~\cite{lam-arXiv-2021} learn the noise scheduling for sampling. The noise schedule for training remains linear as before. After training the score network, they assume it to be close to the optimal value in order to use it for noise schedule training. The inference is composed of two steps. First, the schedule is determined by fixing two initial hyperparameters. The second step is the usual reverse process with the determined schedule. %The authors reported results on CIFAR-10 and CelebA data sets.

% 29. Unleashing Transformers: Parallel Token Prediction with Discrete Absorbing Diffusion for Fast High-Resolution Image Generation from Vector-Quantized Codes
Bond-Taylor \etal~\cite{bond-ECCV-2022} present a two-stage process, where they apply vector quantization to images to obtain discrete representations, and use a transformer \cite{Vaswani-NIPS-2017} to reverse a discrete diffusion process, where the elements are randomly masked at each step. The sampling process is faster because the diffusion is applied to a highly compressed representation, which allows fewer denoising steps (50-256). %This procedure is tested on image generation using the LSUN and FFHQ data sets.The architecture employed is the Transformer of Vaswani \etal~\cite{Vaswani-NIPS-2017}.

% 34.Learning to Efficiently Sample from Diffusion Probabilistic Models
Watson \etal~\cite{watson-arXiv-2021} propose a dynamic programming algorithm that finds the optimal inference schedule, having a time complexity of $\mathcal{O}(T)$, where $T$ is the number of steps. They conduct their image generation experiments on CIFAR-10 and ImageNet, using the DDPM architecture.

% 69\textbf{Learning Fast Samplers for Diffusion Models by Differentiating Through Sample Quality} 
In a different work, Watson \etal~\cite{watson-ICLR-2021} begin by presenting how a reparametrization trick can be integrated within the backward process of diffusion models in order to optimize a family of fast samplers. Using the Kernel Inception Distance as loss function, they show how optimization can be done using stochastic gradient descent. Next, they propose a special parametrized family of samplers, which, using the same process as before, can achieve competitive results with fewer sampling steps. Using FID and Inception Score (IS) as metrics, the method seems to outperform some diffusion model baselines. 

% 59\textbf{TACKLING THE GENERATIVE LEARNING TRILEMMA WITH DENOISING DIFFUSION GANS} 
Similar to Bond-Taylor \etal~\cite{bond-ECCV-2022} and Watson \etal~\cite{watson-arXiv-2021,watson-ICLR-2021}, Xiao \etal~\cite{xiao-arXiv-2021} try to improve the sampling speed, while also maintaining the quality, coverage and diversity of the samples. Their approach is to integrate a GAN in the denoising process to discriminate between real samples (forward process) and fake ones (denoised samples from the generator), with the objective of minimizing the softened reverse KL divergence \cite{shannon-arXiv-2020}. However, the model is modified by directly generating a clean (fully denoised) sample and conditioning the fake example on it. Using the NCSN++ architecture with adaptive group normalization layers for the GAN generator, they achieve similar FID values in both image synthesis and stroke-based image generation, at sampling rates of about 20 to 2000 times faster than other diffusion models. 

% 55. Variational Diffusion Models
Kingma \etal~\cite{kingma-NeurIPS-2021} introduce a class of diffusion models that obtains state-of-the-art likelihoods on image density estimation. They add Fourier features to the input of the network to predict the noise, and investigate if the observed improvement is specific to this class of models. Their results confirm the hypothesis, \ie~previous state-of-the-art models did not benefit from this change. As a theoretical contribution, they show that the diffusion loss is impacted by the signal-to-noise ratio function only through its extremes. %The data sets used in their experiments are LSUN and ImageNet and the architecture employed is DDPM.

% 56\textbf{Analytic-DPM: an Analytic Estimate of the Optimal Reverse Variance in Diffusion Probabilistic Models}. 
Following the work presented in \cite{gao-ICLR-2021}, Bao \etal~\cite{bao-arXiv-2022} propose an inference framework that does not require training using non-Markovian diffusion processes. By first deriving an analytical estimate of the optimal mean and variance with respect to a score function, and using a pretrained scored-based model to obtain score values, they show better results, while being 20 to 40 times more time-efficient. The score is approximated by Monte Carlo sampling. However, the score is clipped within some precomputed bounds in order to diminish any bias of the pretrained DDPM model. % Experiments on image generation were carried out using other’s work as the score-based models which were trained on CelebA, ImageNet and LSUN Bedroom, as well as training their own models on CIFAR-10.

% 71\textbf{Truncated Diffusion Probabilistic Models}. 
Zheng \etal~\cite{zheng-arXiv-2022} suggest truncating the process at an arbitrary step, and propose a method to inverse the diffusion from this distribution by relaxing the constraint of having Gaussian random noise as the final output of the forward diffusion. To solve the issue of starting the reverse process from a non-tractable distribution, an implicit generative distribution is used to match the distribution of the diffused data. The proxy distribution is fit either through a GAN or a conditional transport. We note that the generator utilizes the same U-Net model as the sampler of the diffusion model, thus not adding extra parameters to be trained. % By first evaluating on synthetic data (Swiss Roll, Double Moons, 8-modal, and 25-modal Gaussian mixtures with equal component weights) and then on real data sets (CIFAR-10, CelebA, CelebA-HQ, LSUN bedroom and LSUN Church), the image synthesis quality and diversity are improved while being more time-efficient.

% 82\textbf{On Analyzing Generative and Denoising Capabilities of Diffusion-based Deep Generative Models}. 
Deja \etal~\cite{deja-arXiv-2022} begin by analyzing the backward process of a diffusion model and postulate that it is formed of two models, a generator and a denoiser. Thus, they propose to explicitly split the process into two components: the denoiser via an auto-encoder, and the generator via a diffusion model. Both models use the same U-Net architecture. % The experiments are carried out on FashionMNIST, CIFAR-10, and CelebA data sets, validating the method as well as showing promising results even compared to SOTA diffusion-based models.

% 85\textbf{Diffusion-GAN: Training GANs with Diffusion} 
Wang \etal~\cite{wang-arXiv-2022a} start from the idea presented by Arjovsky \etal~\cite{arjovsky-arXiv-2017} and S{\o}nderby \etal~\cite{sonderby-arXiv-2016} to augment the input data of the discriminator by adding noise. This is achieved in \cite{wang-arXiv-2022a} by injecting noise from a Gaussian mixture distribution composed of weighted diffused samples from the clean image at various time steps. The noise injection mechanism is applied to both real and fake images. The experiments are conducted on a wide range of data sets covering multiple resolutions and high diversity. %: CIFAR-10, STL-10, LSUN-Bedroom, LSUN-Church, AFHQ(Cat/Dog/Wild) and FFHQ. The architectures of the GAN model are StyleGAN2 or ProjectedGAN. Measuring the FID score for synthesis quality and improved recall \cite{kynkaanniemi-NeurIPS-2019} for diversity, the method presents results comparable to state-of-the-art. 

\subsubsection{Score-Based Generative Models}
% 4.Improved Techniques for Training Score-Based Generative Models
Starting from a previous work \cite{song-NeurIPS-2019}, Song \etal~\cite{song-NeurIPS-2020} present several improvements which are based on theoretical and empirical analyses. They address both training and sampling phases. For training, the authors show new strategies to choose the noise scales and how to incorporate the noise conditioning into NCSNs \cite{song-NeurIPS-2019}. For sampling, they propose to apply an exponential moving average to the parameters and select the hyperparameters for the Langevin dynamics such that the step size verifies a certain equation. The proposed changes unlock the application of NCSNs on high-resolution images. %The experiments were conducted on the following data sets: CIFAR-10, CelebA, LSUN and FFHQ.

% 5.Adversarial score matching and improved sampling for image generation
Jolicoeur-Martineau \etal~\cite{jolicoeur-ICLR-2021} introduce an adversarial objective along with denoising score matching to train score-based models. Furthermore, they propose a new sampling procedure called Consistent Annealed Sampling and prove that it is more stable than the annealed Langevin method. Their image generation experiments %on CIFAR-10, LSUN-churches and StackedMNIST
show that the new objective returns higher quality examples without an impact on diversity. The suggested changes are tested on the architectures proposed in \cite{song-NeurIPS-2019, song-NeurIPS-2020, ho-NeurIPS-2020}.

% 9.Maximum Likelihood Training of Score-Based Diffusion Models
Song \etal~\cite{song-NeurIPS-2021} improve the likelihood of score-based diffusion models. They achieve this through a new weighting function for the combination of the score matching losses. For their image generation experiments, they use the DDPM++ architecture introduced in \cite{song-ICLR-2021}.

%14. Diffusion Schrödinger Bridge with Applications to Score-Based Generative Modeling 
In \cite{de-NeurIPS-2021}, the authors present a score-based generative model as an implementation of Iterative Proportional Fitting (IPF), a technique used to solve the Schr{\"o}dinger bridge problem. This novel approach is tested on image generation, as well as data set interpolation, which is possible because the prior can be any distribution.

% 19. Score-based Generative Modeling in Latent Space
Vahdat \etal~\cite{vahdat-NeurIPS-2021} train diffusion models on latent representations. They use a VAE to encode to and decode from the latent space. This work achieves up to 56 times faster sampling. For the image generation experiments, the authors employ the NCSN++ architecture introduced in \cite{song-ICLR-2021}.

\subsubsection{Stochastic Differential Equations}
%13. Diffusion Normalizing Flow 
DiffFlow is introduced in \cite{zhang-NeurIPS-2021} as a new generative modeling approach that combines normalizing flows and diffusion probabilistic models. From the perspective of diffusion models, the method has a sampling procedure that is up to 20 times more efficient, thanks to a learnable forward process which skips unneeded noise regions. The authors perform experiments using the same architecture as in \cite{ho-NeurIPS-2020}.

% 33. Gotta Go Fast When Generating Data with Score-Based Models
Jolicoeur-Martineau \etal~\cite{jolicoeur-arXiv-2021} introduce a new SDE solver that is between $2\times$ and $5\times$ faster than Euler-Maruyama and does not affect the quality of the generated images. The solver is evaluated in a set of image generation experiments with pretrained models from \cite{song-NeurIPS-2019}.

% 35. Deep Generative Learning via Schrödinger Bridge
Wang \etal~\cite{wang-ICML-2021} present a new deep generative model based on Schr{\"o}dinger bridge. This is a two-stage method, where the first stage learns a smoothed version of the target distribution, and the second stage derives the actual target. %CIFAR-10 and ImageNet are the data sets used in their experiments and the architecture is the same as in \cite{song-ICLR-2021b}.

%58\textbf{SCORE-BASED GENERATIVE MODELING WITH CRITICALLY-DAMPED LANGEVIN DIFFUSION}. 
Focusing on scored-based models, Dockhorn \etal~\cite{dockhorn-arXiv-2021} utilize a critically-damped Langevin diffusion process by adding another variable (velocity) to the data, which is the only source of noise in the process. Given the new diffusion space, the resulting score function is demonstrated to be easier to learn. The authors extend their work by developing a more suitable score objective called hybrid score matching, as well as a sampling method, by solving the SDE through integration. The authors adapt the NCSN++ and DDPM++ architectures to accept both data and velocity, being evaluated on unconditional image generation and outperforming similar score-based diffusion models.

% 67\textbf{Heavy-tailed denoising score matching}. 
Motivated by the limitations of high-dimensional score-based diffusion models due to the Gaussian noise distribution, Deasy \etal~\cite{deasy-arXiv-2021} extend denoising score matching to generalize to the normal noising distribution. By adding a heavier tailed distribution, their experiments on several data sets show promising results, as the generative performance improves in certain cases (depending on the shape of the distribution). An important scenario in which the method excels is on data sets with unbalanced classes.

% 76\textbf{Subspace Diffusion Generative Models} 
Jing \etal~\cite{jing-arXiv-2022} try to shorten the duration of the sampling process of diffusion models by reducing the space onto which diffusion is realized, \ie~the larger the time step in the diffusion process, the smaller the subspace. The data is projected onto a finite set of subspaces, at specific times, each being associated with a score model. This results in reduced computational costs, while the performance is increased. The work is limited to natural image synthesis. Evaluating the method in unconditional image generation, the authors achieve similar or better performance compared with state-of-the-art models, while having a lower inference time. The method is demonstrated to work for the inpainting task as well.

% 79\textbf{Maximum Likelihood Training of Implicit Nonlinear Diffusion Models} 
Kim \etal~\cite{kim-arXiv-2022} propose to change the diffusion process into a non-linear one. This is achieved by using a trainable normalizing flow model which encodes the image in the latent space, where it can now be linearly diffused to the noise distribution. %Then, this whole process is projected back to the data space. 
A similar logic is then applied to the denoising process. This method is applied on the NCSN++ and DDPM++ frameworks, while the normalizing flow model is based on ResNet. %Reporting negative log-likelihood, negative evidence lowerbound and FID as metrics, the method shows competitive results in CIFAR-10 and CelebA while outperforming on FID score for CelebA. 

% 86\textbf{Accelerating Score-based Generative Models for High-Resolution Image Synthesis}. 
Ma \etal~\cite{ma-arXiv-2022} aim to make the backward diffusion process more time-efficient, while maintaining the synthesis performance. Within the family of score-based diffusion models, they begin to analyze the reverse diffusion in the frequency domain, subsequently applying a space-frequency filter to the sampling process, which aims to integrate information about the target distribution into the initial noise sampling. The authors conduct experiments with NCSN \cite{song-NeurIPS-2019} and NCSN++ \cite{song-ICLR-2021}, where the proposed method clearly shows speed improvements in image synthesis (by up to 20 times less sampling steps), while keeping the same satisfactory generation quality for both low and high-resolution images.

\subsection{Conditional Image Generation}

We next showcase diffusion models that are applied to conditional image synthesis. The condition is commonly based on various source signals, in most cases some class labels being used. Some methods perform both unconditional and conditional generation, which are also discussed here. 

\subsubsection{Denoising Diffusion Probabilistic Models}

%11. Diffusion Models Beat GANs on Image Synthesis 
Dhariwal \etal~\cite{dhariwal-NeurIPS-2021} introduce few architectural changes to improve the FID of diffusion models. They also propose classifier guidance, a strategy which uses the gradients of a classifier to guide the diffusion during sampling. They conduct both unconditional and conditional image generation experiments.

% 66image-cond\textbf{High Fidelity Visualization of What Your Self-Supervised Representation Knows About}. 
Bordes \etal~\cite{bordes-arXiv-2021} examine representations resulting from self-supervised tasks by visualizing and comparing them to the original image. They also compare representations generated from different sources. Thus, a diffusion model is used to generate samples that are conditioned on these representations. The authors implement several modifications to the U-Net architecture presented by Dhariwal \etal~\cite{dhariwal-NeurIPS-2021}, such as adding conditional batch normalization layers, and mapping the vector representation through a fully connected layer. % The experiments were realized on ImageNet, conditioning on representations generated from multiple methods (VicReg, Dino, Barlow Twins and SimCLR), as well as representations from the projector of the methods (except VicReg). The results are competitive with other previous methods. 

% 45. Generating High Fidelity Data from Low-density Regions using Diffusion Models 
The method presented in \cite{sehwag-CVPR-2022} allows diffusion models to produce images from low-density regions of the data manifold. They use two new losses to guide the reverse process. The first loss guides the diffusion towards low-density regions, while the second enforces the diffusion to stay on the manifold. Moreover, they demonstrate that their diffusion model does not memorize the examples from the low-density neighborhoods, generating novel images. The authors employ an architecture similar to that of Dhariwal \etal~\cite{dhariwal-NeurIPS-2021}.

% 25.On Fast Sampling of Diffusion Probabilistic Models
Kong \etal~\cite{kong-ICMLW-2021} define a bijection between the continuous diffusion steps and the noise levels. With the defined bijection, they are able to construct an approximate diffusion process which requires less steps. The method is tested using the previous DDIM \cite{song-ICLR-2021b} and DDPM \cite{ho-NeurIPS-2020} architectures on image generation. % with LSUN, CelebA and CIFAR-10 data sets.

% 38. DiffuseVAE: Efficient, Controllable and High-Fidelity Generation from Low-Dimensional Latents
Pandey \etal~\cite{pandey-NeurIPSW-2021} build a generator-refiner framework, where the generator is a VAE and the refiner is a DDPM conditioned by the output of the VAE. The latent space of the VAE can be used to control the content of the generated image because the DDPM only adds the details. After training the framework, the resulting DDPM is able to generalize to different noise types. More specifically, if the reverse process is not conditioned on the VAE's output, but on different noise types, the DDPM is able to reconstruct the initial image. %The experiments are conducted on CelebA-HQ 64x64,128x128 and CIFAR-10 using DDPM.

% 53. Cascaded Diffusion Models for High Fidelity Image Generation
Ho \etal~\cite{ho-arXiv-2021} introduce Cascaded Diffusion Models (CDM), an approach for generating high-resolution images conditioned on ImageNet classes. Their framework contains multiple diffusion models, where the first model from the pipeline generates low-resolution images conditioned on the image class. The subsequent models are responsible for generating images of increasingly higher resolutions. These models are conditioned on both the class and the low-resolution image. %The data sets used in their experiments are the LSUN and ImageNet data sets and as diffusion models they use DDPM.

% 42. Dynamic Dual-Output Diffusion Models
Benny \etal~\cite{benny-CVPR-2022} study the advantages and disadvantages of predicting the image instead of the noise during the reverse process. They conclude that some of the discovered problems could be addressed by interpolating the two types of output. They modify previous architectures to return both the noise and the image, as well as a value that controls the importance of the noise when performing the interpolation. The strategy is evaluated on top of the DDPM and DDIM architectures. %CIFAR-10, ImageNet and CelebA and the diffusion models used are DDPM and DDIM.

% 48. Perception Prioritized Training of Diffusion Models 
Choi \etal~\cite{choi-CVPR-2022} investigate the impact of the noise levels on the visual concepts learned by diffusion models. They modify the conventional weighting scheme of the objective function to a new one that enforces diffusion models to learn rich visual concepts. The method groups the noise levels into three categories (coarse, content and clean-up) according to the signal-to-noise ratio, \ie~small SNR is coarse, medium SNR is content, large SNR is clean-up. The weighting function assigns lower weights to the last group. %They use DDPM in their experiments and FFHQ, AFHQ-D, CUB, Flowers, MetFaces as data sets.

% 77noise-cond\textbf{On Conditioning the Input Noise for Controlled Image Generation with Diffusion Models}
Singh \etal~\cite{singh-arXiv-2022} propose a novel method for conditional image generation. Instead of conditioning the signal throughout the sampling process, they present a method to condition the noise signal (from where the sampling starts). Using Inverting Gradients \cite{geiping-NeurIPS-2020}, the noise is injected with information about localization and orientation of the conditioned class, while maintaining the same random Gaussian distribution. % Experiments on ImageNet were carried out, generating samples of different classes and using various input noises while varying the number of steps in Inverting Gradients. 

% 106text-cond+others-cond\textbf{Compositional Visual Generation with Composable Diffusion Models}. 
Describing the resembling functionality of diffusion models and energy-based models, and leveraging the compositional structure of the latter models, Liu \etal~\cite{liu-arXiv-2022b} propose to combine multiple diffusion models for conditional image synthesis. In the reverse process, the composition of multiple diffusion models, each associated with a different condition, can be achieved either through conjunction or negation. % The method was evaluated on CLEVR, Relational CLEVR and FFHQ and compared to EBM \cite{du-NeurIPS-2020}, StyleGAN2 \cite{karras-CVPR-2020}, LACE \cite{nie-NeurIPS-2021} and GLIDE \cite{nichol-arXiv-2021} baselines for image synthesis using various conditions: text, 2D positions, relational descriptions between objects or human facial attributes. Reporting the Clean-FID \cite{parmar-CVPR-2022}, the model significantly outperforms the baselines.

\subsubsection{Score-Based Generative Models}

% 73class-cond\textbf{Denoising Likelihood Score Matching for Conditional Score-based Data Generation}. 
The works of Song \etal~\cite{song-ICLR-2021} and Dhariwal \etal~\cite{dhariwal-NeurIPS-2021} on scored-based conditional diffusion models based on classifier guidance inspired Chao \etal~\cite{chao-arXiv-2022} to develop a new training objective which reduces the potential discrepancy between the score model and the true score. The loss of the classifier is modified into a scaled cross-entropy added to a modified score matching loss. % Evaluating on the CIFAR-10 and CIFAR-100 baselines, it outperforms the metrics that measure the sample quality (IS and FID), as well as metrics assessing how genuine class samples are (Precision, Recall, Density, Coverage for each individual class).

\subsubsection{Stochastic Differential Equations}

% 24. Classifier-Free Diffusion Guidance
Ho \etal~\cite{ho-NeurIPS-2021} introduce a guidance method that does not require a classifier. It just needs one conditional diffusion model and one unconditional version, but they use the same model to learn both cases. The unconditional model is trained with the class identifier being equal to 0. The idea is based on the implicit classifier derived from the Bayes rule. %The data sets used for image generation are ImageNet 64x64 and 128x128.
%2021

% 39. Pseudo Numerical Methods for Diffusion Models on Manifolds
Liu \etal~\cite{liu-ICLR-2022} investigate the usage of conventional numerical methods to solve the ODE formulation of the reverse process. They find that these methods return lower quality samples compared with the previous approaches. Therefore, they introduce pseudo-numerical methods for diffusion models. Their idea splits the numerical methods into two parts, the gradient part and the transfer part. The transfer part (standard methods have a linear transfer part) is replaced such that the result is as close as possible to the target manifold. As a last step, they show how this change solves the problems discovered when using conventional approaches. %The experiments are conducted on the CIFAR-10 and CelebA data sets.

% 68uncond+cond-speech\textbf{Itô-Taylor Sampling Scheme for Denoising Diffusion Probabilistic Models using Ideal Derivatives} 
Tachibana \etal~\cite{tachibana-arXiv-2021} address the slow sampling problem of DDPMs. They propose to decrease the number of sampling steps by increasing the order (from one to two) of the stochastic differential equation solver (denoising part). While preserving the network architecture and score matching function, they adopt the It\^{o}-Taylor expansion scheme for the sampler, as well as substitute some derivative terms in order to simplify the calculation. They reduce the number of backward steps while retaining the performance. On top of these, another contribution is the new noise schedule. %The first type of evaluation is done on unconditional image synthesis, showing that it obtains a better FID score on CelebA data set in fewer steps, while the overall performance for greater steps is slightly better. The second evaluation is on conditional speech synthesis, reporting less noisy and clearer overtones on LJSpeech data set compared to WaveGrad by Vovk \url{https://github.com/ivanvovk/WaveGrad}. 

% 57uncond+class-cond\textbf{Elucidating the Design Space of Diffusion-Based Generative Models}
Karras \etal~\cite{karras-arXiv-2022} try to separate diffusion scored-based models into individual components that are independent of each other. This separation allows modifying a single part without affecting the other units, thus facilitating the improvement of diffusion models. Using this framework, the authors first present a sampling process that uses Heun's method as the ODE solver, which reduces the neural function evaluations while maintaining the FID score. They further show that a stochastic sampling process brings great performance benefits. The second contribution is related to training the score-based model by preconditioning the neural network on its input and the corresponding targets, as well as using image augmentation. % The experiments consisted of both unconditional (CIFAR-10) and conditional (ImageNet-64) image generation tasks. 

% 60uncond+class-cond\textbf{PROGRESSIVE DISTILLATION FOR FAST SAMPLING OF DIFFUSION MODELS}. 
Within the context of both unconditional and class-conditional image generation, Salimans \etal~\cite{salimans-arXiv-2022} propose a technique for reducing the number of sampling steps. They distill the knowledge of a trained teacher model, represented by a deterministic DDIM, into a student model that has the same architecture, but halving the number of sampling steps. In other words, the target of the student is to take two consecutive steps of the teacher. Furthermore, this process can be repeated until the desired number of sampling steps is reached, while maintaining the same image synthesis quality. Finally, three versions of the model and two loss functions are explored in order to facilitate the distillation process and reduce the number of sampling steps (from 8192 to 4). %The experiments on image generation on various data sets (CIFAR-10, ImageNet, and LSUN) show that sampling steps can be reduced from 8192 steps to 4 while keeping a similar FID score.

% 81uncond+cond-music\textbf{A Continuous Time Framework for Discrete Denoising Models} 
Campbell \etal~\cite{campbell-arXiv-2022} demonstrate a continuous-time formulation of denoising diffusion models that is capable of working with discrete data. The work models the forward continuous-time Markov chain diffusion process via a transition rate matrix, and the backward denoising process via a parametric approximation of the inverse transition rate matrix. Further contributions are related to the training objective, the matrix construction, and an optimized sampler. %The method is firstly validated on 2d synthetic data, and then it is evaluated on CIFAR-10 for image synthesis and on Lakh pianoroll data set for music generation, having better results than the discre-time methods. 

% 84uncond+class-cond\textbf{DPM-Solver: A Fast ODE Solver for Diffusion Probabilistic Model Sampling in Around 10 Steps}.
The interpretation of diffusion models as ODEs proposed by Song \etal~\cite{song-ICLR-2021} is reformulated by Lu \etal~\cite{lu-arXiv-2022} in a form that can be solved using an exponential integrator. Other contributions of Lu \etal~\cite{lu-arXiv-2022} are an ODE solver that approximates the integral term of the new formulation using Taylor expansion (first order to third order), and an algorithm that adapts the time step schedule, being 4 to 16 times faster. % Evaluating on CIFAR-10, CelebA, ImageNet and LSUN bedroom data sets and comparing with other diffusion models on image generation that have similar lower sampling steps, it produced good results while being 4 to 16 times faster.

\subsection{Image-to-Image Translation}

%2021
% 61\textbf{Palette: Image-to-Image Diffusion Models} 
Saharia \etal~\cite{saharia-SIGGRAPH-2022} propose a framework for image-to-image translation using diffusion models, focusing on four tasks: colorization, inpainting, uncropping and JPEG restoration. The proposed framework is the same across all four tasks, meaning that it does not suffer custom changes for each task. The authors begin by comparing $L_1$ and $L_2$ losses, suggesting that $L_2$ is preferred, as it leads to a higher sample diversity. %For evaluation, they propose two benchmark data sets: ctest10k \cite{larsson-ECCV-2016} - a subset from ImageNet validation set and places10k \cite{zhou-PAMI-2017} - a subset from Places2 validation set, and how to report automated metrics on these in order to create a unified evaluation protocol. 
Finally, they reconfirm the importance of self-attention layers in conditional image synthesis. 

% 101\textbf{UNIT-DDPM: UNpaired Image Translation with Denoising Diffusion Probabilistic Models}. 
To translate an unpaired set of images, Sasaki \etal~\cite{sasaki-arXiv-2021} propose a method involving two jointly trained diffusion models. During the reverse denoising process, at every step, each model is also conditioned on the other's intermediate sample. Furthermore, the loss function of the diffusion models is regularized using the cycle-consistency loss \cite{zhu-ICCV-2017}. % The architectures of the models consist of U-Nets that integrate features from PixelCNN \cite{salimans-arXiv-2017} and Wide ResNet \cite{zagoruyko-arXiv-2016}. The CycleGAN \cite{zhu-ICCV-2017}, UNIT \cite{liu-NeurIPS-2017}, MUNIT \cite{huang-ECCV-2018} and DRIT++ \cite{lee-ECCV-2018} were used as baselines for comparison. The results on CMP Facades AND KAIST Multispectral Pedestrian data sets, as well as images scraped from google maps from \cite{zhu-ICCV-2017}, showed the method outperforms by 20\%. 

%2022
% 94\textbf{EGSDE: Unpaired Image-to-Image Translation via Energy-Guided Stochastic Differential Equations}. 
The aim of Zhao \etal~\cite{zhao-arXiv-2022} is to improve current image-to-image translation score-based diffusion models by utilizing data from a source domain with an equal significance. An energy-based function trained on both source and target domains is employed in order to guide the SDE solver. This leads to generating images that preserve the domain-agnostic features, while translating characteristics specific to the source domain to the target domain. The energy function is based on two feature extractors, each specific to a domain. % The data sets used in the evaluations are CelebaA-HQ (translating man to female) and AFHQ (for cat to dog and wild to dog). The model reports superior performance on all metrics compared to the extensive list of baselines (both GANs and diffusion models). 

% 95\textbf{Pretraining is All You Need for Image-to-Image Translation}. 
Leveraging the power of pretraining, Wang \etal~\cite{wang-arXiv-2022c} employ the GLIDE model \cite{nichol-arXiv-2021} and train it to obtain a rich semantic latent space. Starting from the pretrained version and replacing the head to adapt to any conditional input, the model is fine-tuned on some specific image generation downstream tasks. This is done in two steps, where the first step is to freeze the decoder and train only the new encoder, and the second step is to train them simultaneously. Finally, the authors employ adversarial training and normalize the classifier-free guidance to enhance generation quality. % The evaluation is carried out of three data sets, each with a different image-to-image translation task: ADE20K (mask-to-image), COCO-Stuff (sketch-to-image), DIODE (geometry-to-image); and compared to Pix2PixHD, SPADE and OASIS baselines. Reporting the FID score, the proposed method significantly outperforms the state of the art. 

% 96\textbf{VQBB: Image-to-image Translation with Vector Quantized Brownian Bridge} 
Li \etal~\cite{li-arXiv-2022} introduce a diffusion model for image-to-image translation that is based on Brownian bridges, as well as GANs. The proposed process begins by encoding the image with a VQ-GAN \cite{esser-CVPR-2021}. Within the resulting quantized latent space, the diffusion process, formulated as a Brownian bridge, maps between the latent representations of the source domain and target domain. Finally, another VQ-GAN decodes the quantized vectors in order to synthesize the image in the new domain. The two GAN models are independently trained on their specific domains. %The diffusion model based on Brownian bridges adopts the same U-Net architecture as described in [3]. The data sets used in the experiments to validate the method are face2comic, edges2shoes and edges2handbags. 

% 98\textbf{The Swiss Army Knife for Image-to-Image Translation: Multi-Task Diffusion Models}. 
Continuing their previous work proposed in \cite{wolleb-arXiv-2022a}, Wolleb \etal~\cite{wolleb-arXiv-2022b} extend the diffusion model by replacing the classifier with another model specific to the task. Thus, at every step of the sampling process, the gradient of the task-specific network is infused. The method is demonstrated with a regressor (based on an encoder) or with a segmentation model (using the U-Net architecture), whereas the diffusion model is based on existing frameworks \cite{ho-NeurIPS-2020,nichol-ICML-2021}. This setting has the advantage of eliminating the need to retrain the whole diffusion model, except for the task-specific model. % Experiments were conducted on a data set containing faces of people and their corresponding age (translate the face to a given age), as well as on BRATS2020 (translate the healthy brain to exhibit tumor). The satisfying results confirm the competence of the method.

\subsection{Text-to-Image Synthesis}

Perhaps the most impressive results of diffusion models are attained on text-to-image synthesis, where the capability of combining unrelated concepts, such as objects, shapes and textures, to generate unusual examples comes to light. To confirm this statement, we used Stable Diffusion \cite{rombach-CVPR-2022} to generate images based on various text prompts, and the results are shown in Figure~\ref{fig_tti_examples}.

% \begin{figure}[!t]
% \begin{center}
% \centerline{\includegraphics[width=0.3\linewidth]{moon_rabbit.png}
% \includegraphics[width=0.3\linewidth]{moon_rabbit_2.png}
% \includegraphics[width=0.3\linewidth]{moon_rabbit_3.png}}
% \vspace{-0.25cm}
% \caption{Pictures of ``a stone rabbit statue sitting on the moon'' generated by Stable Diffusion \cite{rombach-CVPR-2022} via the \url{https://beta.dreamstudio.ai/dream} platform.}
% \label{fig_example}
% \vspace{-0.4cm}
% \end{center}
% \end{figure}

%54 Photorealistic Text-to-Image Diffusion Models with Deep Language Understanding
Imagen is introduced in \cite{saharia-arXiv-2022} as an approach for text-to-image synthesis. It consists of one encoder for the text sequence and a cascade of diffusion models for generating high-resolution images. These models are also conditioned on the text embeddings returned by the encoder. Moreover, the authors introduce a new set of captions (DrawBench) for text-to-image evaluations. %They used it for human evaluation. 
Regarding the architecture, the authors develop Efficient U-Net to improve efficiency, and apply this architecture in their text-to-image generation experiments.

%2021
% 50. Vector Quantized Diffusion Model for Text-to-Image Synthesis 
Gu \etal~\cite{gu-CVPR-2022} introduce the VQ-Diffusion model, a method for text-to-image synthesis that does not have the unidirectional bias of previous approaches. With its masking mechanism, the proposed method avoids the accumulation of errors during inference. The model has two stages, where the first stage is based on a VQ-VAE that learns to represent an image via discrete tokens, and the second stage is a discrete diffusion model that operates on the discrete latent space of the VQ-VAE. The training of the diffusion model is conditioned on caption embeddings. Inspired from masked language modeling, some tokens are replaced with a \emph{[mask]} token. %The experiments are conducted on text-to-image synthesis using CUB-200, Oxford-102 and MSCOCO data sets.

%2022
% 52. Hierarchical Text-Conditional Image Generation with CLIP Latents
Avrahami \etal~\cite{avrahami-CVPR-2022} present a text-conditional diffusion model conditioned on CLIP \cite{radford-ICML-2021} image and text embeddings. This is a two-stage approach, where the first stage generates the image embedding, and the second stage (decoder) produces the final image conditioned on the image embedding and the text caption. To generate image embeddings, the authors use a diffusion model in the latent space. They perform a subjective human assessment to evaluate their generative results.

% 75uncond+text-cond\textbf{Fast Sampling of Diffusion Models with Exponential Integrator}. 
Addressing the slow sampling inconvenience of diffusion models, Zhang \etal~\cite{zhang-arXiv-2022} focus their work on a new discretization scheme that reduces the error and allows a greater step size, \ie~a lower number of sampling steps. By using high-order polynomial extrapolations in the score function and an Exponential Integrator for solving the reverse SDE, the number of network evaluations is drastically reduced, while maintaining the generation capabilities. %This is shown in the evaluation on CIFAR-10, CELEBA and ImageNet, outperforming baselines that require a similar small number of sampling steps (DDIM, A-DDIM and PNDM).

% 104uncond+text-cond\textbf{DiVAE: Photorealistic Images Synthesis with Denoising Diffusion Decoder} 
Shi \etal~\cite{shi-arXiv-2022b} combine a VQ-VAE \cite{van-NIPS-2017} and a diffusion model to generate images. Starting from the VQ-VAE, the encoding functionality is preserved, while the decoder is replaced by a diffusion model. The authors use the U-Net architecture from \cite{nichol-ICML-2021}, injecting the image tokens into the middle block. %Firstly, the model is compared to state-of-the-art models (VQ-VAE, VQ-GAN) on the validation set of ImageNet for unconditional image synthesis, showing better FID scores. The model is also evaluated for conditional image generation (text-to-image) with an autoregressive generator pretrained on Conceptual Captions and tested on MSCOCO, achieving the lowest FID excluding GLIDE and DALLE-2. 

% 105text-cond\textbf{Text-Guided Synthesis of Artistic Images with Retrieval-Augmented Diffusion Models}. 
Building on top of the work presented in \cite{blattmann-arXiv-2022}, Rombach \etal~\cite{rombach-arXiv-2022} introduce a modification to create artistic images using the same procedure: extract the k-nearest neighbors in the CLIP \cite{radford-ICML-2021} latent space of an image from a database, then generate a new image by guiding the reverse denoising process with these embeddings. As the CLIP latent space is shared by text and images, the diffusion can be guided by text prompts as well. However, at inference time, the database is replaced with another one that contains artistic images. Thus, the model generates images within the style of the new database. %Two experiments were done: firstly training the model on  OpenImages and then using WikiAr for the stylised database and secondly LAION-2B-en with ArtBench, results showing an increase in performance.

% 107text-cond\textbf{Text2Human: Text-Driven Controllable Human Image Generation} 
Jiang \etal~\cite{jiang-arXiv-2022} present a framework to generate images of full-body humans with rich clothing representation given three inputs: a human pose, a text description of the clothes' shape, and another text of the clothing texture. The first stage of the method encodes the former text prompt into an embedding vector and infuses it into the module (encoder-decoder based) that generates a map of forms. In the second stage, a diffusion-based transformer samples an embedded representation of the latter text prompt from multiple multi-level codebooks (each specific to a texture), a mechanism suggested in VQ-VAE \cite{van-NIPS-2017}. Initially, the codebook indices at coarser levels are sampled, and then, using a feed-forward network, the finer-level indices are predicted. The text is encoded using Sentence-BERT \cite{reimers-arXiv-2019}. % In order to evaluate their approach, the DeepFashionMultiModal data set was created from DeepFashion data set, containing human poses, the masking map of forms and annotations describing clothes with their shape and texture. When compared to other models (Pix2PixHD \cite{wang-CVPR-2018}, SPADE \cite{Park-CVPR-2019}, MISC \cite{Weng-CVPR-2020}, HumanGAN \cite{sarkar-3DV-2021}, TryOnGAN \cite{lewis-TOG-2021}, Taming Transformer \cite{esser-CVPR-2021}), the method obtains the lowest FID score.

\subsection{Image Super-Resolution}
% 20.Image Super-Resolution via Iterative Refinement
Saharia \etal~\cite{saharia-arXiv-2021} apply diffusion models to super-resolution. Their reverse process learns to generate high quality images conditioned on low-resolution versions. This work employs the architectures presented in \cite{ho-NeurIPS-2020, nichol-ICML-2021} and the following data sets: CelebA-HQ, FFHQ and ImageNet.

%  36.Score-based Generative Neural Networks for Large-Scale Optimal Transport
Daniels \etal~\cite{daniels-NeurIPS-2021} use score-based models to sample from the Sinkhorn coupling of two distributions. Their method models the dual variables with neural networks, then solves the problem of optimal transport. After training the neural networks, the sampling can be performed via Langevin dynamics and a score-based model. They run experiments on image super-resolution using a U-Net architecture.

\subsection{Image Editing}
%2021
% 63\textbf{SDEdit: Guided Image Synthesis and Editing with Stochastic Differential Equations} 
Meng \etal~\cite{meng-arXiv-2021} utilize diffusion models in various guided image generation tasks, \ie~stroke painting or stroke-based editing and image composition. Starting from an image that contains some form of guidance, its properties (such as shapes and colors) are preserved, while the deformations are smoothed out by progressively adding noise (forward process of the diffusion model). Then, the result is denoised (reverse process) to create a realistic image according to the guidance. Images are synthesized with a generic diffusion model by solving the reverse SDE, without requiring any custom data set or modifications for training. %The authors present great results, showing better scores in both image realism and faithfulness on LSUN and CelebA-HQ for all three tasks compared to previous GAN baselines. 

% 51. Blended Diffusion for Text-driven Editing of Natural Image
One of the first approaches for editing specific regions of images based on natural language descriptions is introduced in \cite{avrahami-CVPR-2022}. The regions to be modified are specified by the user via a mask. The method relies on CLIP guidance to generate an image according to the text input, but the authors observe that combining the output with the original image at the end does not produce globally coherent images. Hence, they modify the denoising process to fix the issue. More precisely, after each step, the authors apply the mask on the latent image, while also adding the noisy version of the original image. 

%2022
% 103\textbf{Blended Latent Diffusion}. 
Extending the work presented in \cite{rombach-CVPR-2022}, Avrahami \etal~\cite{avrahami-arXiv-2022} apply latent diffusion models for editing images locally, using text. A VAE encodes the image and the adaptive-to-time mask (region to edit) into the latent space where the diffusion process occurs. Each sample is iteratively denoised, while being guided by the text within the region of interest. However, inspired by Blended Diffusion \cite{avrahami-CVPR-2022}, the image is combined with the masked region in the latent space that is noised at the current time step. Finally, the sample is decoded with the VAE to generate the new image. %The method was compared to the following baselines on their associated data sets: Local CLIP-guided diffusion
%\url{https://colab.research.google.com/drive/12a_Wrfi2_gwwAuN3VvMTwVMz9TfqctNj}
%, PaintByWord++ \cite{bau-arXiv-2021}, Blended Diffusion, GLIDE and GLIDE-filtered \cite{nichol-arXiv-2021}. 
The method demonstrates superior performance while being comparably faster.

\subsection{Image Inpainting}

% 17. GLIDE: Towards Photorealistic Image Generation and Editing with Text-Guided Diffusion Models
Nichol \etal~\cite{nichol-arXiv-2021} train a diffusion model conditioned on text descriptions and also study the effectiveness of classifier-free and CLIP-based guidance. They obtain better results with the first option. Moreover, they fine-tune the model for image inpainting, unlocking image modifications based on text input. % The data set used in these experiments was MS-COCO and the architecture was similar with the one from \cite{dhariwal-NeurIPS-2021}.

% 41. RePaint: Inpainting using Denoising Diffusion Probabilistic Models 
Lugmay \etal~\cite{lugmayr-CVPR-2022} present an inpainting method agnostic to the mask form. They use an unconditional diffusion model for this, but modify its reverse process. They produce the image at step $t-1$ by sampling the known region from the masked image, and the unknown region by applying denoising to the image obtained at step $t$. With this procedure, the authors observe that the unknown region has the right structure, while also being semantically incorrect. Further, they solve the issue by repeating the proposed step for a number of times and, at each iteration, they replace the previous image from step $t$ with a new sample obtained from the denoised version generated at step $t-1$. %The method is evaluated on inpainting using CelebA-HQ and ImageNet, while the architecture employed is DDPM.

\subsection{Image Segmentation}
%2021
% 64\textbf{Label-Efficient Semantic Segmentation with Diffusion Models} 
Baranchuk \etal~\cite{baranchuk-arXiv-2021} demonstrate how diffusion models can be used in semantic segmentation. Taking the feature maps (middle blocks) at different scales from the decoder of the U-Net (used in the denoising process) and concatenating them (upsampling the feature maps in order to have the same dimensions), they can be used to classify each pixel by further attaching an ensemble of multi-layer perceptrons. The authors show that these feature maps, extracted at later steps in the denoising process, contain rich representations. %The evaluation was made on two data sets annotated within the research work (based on three classes from LSUN and FFHQ-256), as well as two open-source already-annotated data sets (ADE-Bedroom-30 and CelebA-19). 
The experiments show that segmentation based on diffusion models outperforms most baselines. 

% 90\textbf{SegDiff: Image Segmentation with Diffusion Probabilistic Models} 
Amit \etal~\cite{amit-arXiv-2021} propose the use of diffusion probabilistic models for image segmentation through extending the architecture of the U-Net encoder. The input image and the current estimated image are passed through two different encoders and combined together through summation. The result is then supplied to the encoder-decoder of the U-Net. Due to the stochastic noise infused at every time step, multiple samples for a single input image are generated and used to compute the mean segmentation map. The U-Net architecture is based on a previous work \cite{nichol-ICML-2021}, while the input image generator is built with Residual Dense Blocks \cite{Wang-ECCV-2018}. The denoised sample generator is a simple 2D convolutional layer. % Experiments are done using Cityscapes, Vaihingen and MoNuSeg data sets, reporting results of various metrics (IoU, F1-score, Weighted Coverage or Boundary F-score) and obtaining state-of-the-art performance. 

\subsection{Multi-Task Approaches}

A series of diffusion models have been applied to multiple tasks, demonstrating a good generalization capacity across tasks. We discuss such contributions below.

%2. Generative Modeling by Estimating Gradients of the Data Distribution
Song \etal~\cite{song-NeurIPS-2019} present the noise conditional score network (NCSN), an approach which estimates the score function at different noise scales. For sampling, they introduce an annealed version of Langevin dynamics and use it to report results in image generation and inpainting. %The FID score reported for image generation on CIFAR-10 were comparable with previous top results, while for inpainting the authors highlight several qualitative results. 
The NCSN architecture is mainly based on the work presented in \cite{Lin-CVPR-2017}, with small changes such as replacing batch normalization with instance normalization.

%112. Solving Linear Inverse Problems Using the Prior Implicit in a Denoiser
Kadkhodaie \etal~\cite{kadkhodaie-NeurIPS-2021} train a neural network to restore images corrupted with Gaussian noise, generated using random standard deviations that are restricted to a particular range. After training, the difference between the output of the neural network and the noisy image received as input is proportional with the gradient of the log-density of the noisy data. This property is based on previous work done in \cite{miyasawa-BIIS-1996}. For image generation, the authors use the mentioned difference as gradient (score) estimation and sample from the implicit data prior of the network by employing an iterative method similar to the annealed Langevin dynamics from \cite{song-NeurIPS-2019}. However, the two sampling methods have some dissimilarities, for example the noise injected in the iterative updates follow distinct strategies. In \cite{kadkhodaie-NeurIPS-2021}, the injected noise is adapted according to the network's estimate, while in \cite{song-NeurIPS-2019}, it is fixed. Moreover, the gradient estimates in \cite{song-NeurIPS-2019} are learned by score matching, while Kadkhodaie \etal~\cite{kadkhodaie-NeurIPS-2021} rely on the previously mentioned property to compute the gradients. The contribution of Kadkhodaie \etal~\cite{kadkhodaie-NeurIPS-2021} develops even further by adapting the algorithm to linear inverse problems, such as deblurring and super-resolution.

% 7. Score-Based Generative Modeling through Stochastic Differential Equations
The SDE formulation of diffusion models introduced in \cite{song-ICLR-2021} generalizes over several previous methods \cite{ho-NeurIPS-2020,  sohl-icml-2015, song-NeurIPS-2019}. Song \etal~\cite{song-ICLR-2021} present the forward and reverse diffusion processes as solutions of SDEs. This technique unlocks new sampling methods, such as the Predictor-Corrector sampler, or the deterministic sampler based on ODEs. The authors carry out experiments on image generation, inpainting and colorization. %The data sets used for image generation were CIFAR-10 and CelebA-HQ, while for the other two tasks they used LSUN.

% 92uncond+super-resolution+inpainting+image2image\textbf{Non-Uniform Diffusion Models} 
Batzolis \etal~\cite{batzolis-arXiv-2022} introduce a new forward process in diffusion models, called non-uniform diffusion. This is determined by each pixel being diffused with a different SDE. Multiple networks are employed in this process, each corresponding to a different diffusion scale. The paper further demonstrates a novel conditional sampler that interpolates between two denoising score-based sampling methods. The model, whose architecture is based on \cite{ho-NeurIPS-2020} and \cite{song-ICLR-2021}, is evaluated on unconditional synthesis, super-resolution, inpainting and edge-to-image translation.

% 10. ImageBART: Bidirectional Context with Multinomial Diffusion for Autoregressive Image Synthesis 
Esser \etal~\cite{esser-NeurIPS-2021} propose ImageBART, a generative model which learns to revert a multinomial diffusion process on compact image representations. A transformer is used to model the reverse steps autoregressively, where the encoder's representation is obtained using the output at the previous step. ImageBART is evaluated on unconditional, class-conditional and text-conditional image generation, as well as local editing. % The data sets used in their experiments were LSUN, FFHQ, ImageNet and Conceptual Captions.

% 23. LEARNING ENERGY-BASED MODELS BY DIFFUSION RECOVERY LIKELIHOOD
Gao \etal~\cite{gao-ICLR-2021} introduce diffusion recovery likelihood, a new training procedure for energy-based models. They learn a sequence of energy-based models for the marginal distributions of the diffusion process. Thus, instead of approximating the reverse process with normal distributions, they derive the conditional distributions from the marginal energy-based models. The authors run experiments on both image generation and inpainting. % task using the CelebA and LSUN data sets, while for image generation they employ the CIFAR-10 data set as well. The architecture is based on Wide ResNet.

% 30. Conditional Image Generation with Score-Based Diffusion Models
Batzolis \etal~\cite{batzolis-arXiv-2021} analyze the previous score-based diffusion models on conditional image generation. Moreover, they present a new method for conditional image generation called conditional multi-speed diffusive estimator (CMDE). This method is based on the observation that diffusing the target image and the condition image at the same rate, might be suboptimal. Therefore, they propose to diffuse the two images, which have the same drift and different diffusion rates, with an SDE. The approach is evaluated on inpainting, super-resolution and edge-to-image synthesis.

% 31.More Control for Free! Image Synthesis with Semantic Diffusion Guidance
Liu \etal~\cite{liu-arXiv-2021} introduce a framework which allows text, content and style guidance from a reference image. The core idea is to use the direction that maximizes the similarity between the representations learned for image and text. The image and text embeddings are produced by the CLIP model \cite{radford-ICML-2021}. To address the need of training CLIP on noisy images, the authors present a self-supervised procedure that does not require text captions. The procedure uses pairs of normal and noised images to maximize the similarity between positive pairs and minimize it for negative ones (contrastive objective). % They conduct experiments on FFHQ and LSUN data sets using the DPPM \cite{ho-NeurIPS-2020} architecture.

%2021
% 65image-cond+image2image+editing+\textbf{ILVR: Conditioning Method for Denoising Diffusion Probabilistic Models}
Choi \etal~\cite{choi-arXiv-2021} propose a novel method, which does not require further training, for conditional image synthesis using unconditional diffusion models. Given a reference image, \ie~the condition, each sample is drawn closer to it by eliminating the low frequency content and replacing it with content from the reference image. The low pass filter is represented by a downsampling operation, which is followed by an upsampling filter of the same factor. %, which determines the order of the filter. 
The authors show how this method can be applied on various image-to-image translation tasks, \eg~paint-to-image, and editing with scribbles. %In order to evaluate the quality of samples generated, the FID and LPIPS metrics are reported on FFHQ and METFACES data sets. 

%2022

% 43. Global Context with Discrete Diffusion in Vector Quantised Modelling for Image Generation
Hu \etal~\cite{hu-CVPR-2022} propose to apply diffusion models on discrete representations given by a discrete VAE. They evaluate the idea in image generation and inpainting experiments, considering the CelebA-HQ and LSUN Church data sets. 

% 44.High-Resolution Image Synthesis with Latent Diffusion Models
Rombach \etal~\cite{rombach-CVPR-2022} introduce latent diffusion models, where the forward and reverse processes happen on the latent space learned by an auto-encoder. They also include cross-attention in the architecture, which brings further improvements on conditional image synthesis. The method is tested on super-resolution, image generation and inpainting. % and the data sets used are ImageNet,  CelebA-HQ, FFHQ, LSUN Churches, Bedrooms and Places.

% 46. Diffusion Auto-encoders: Toward a Meaningful and Decodable Representation
The method introduced by Preechakul \etal~\cite{preechakul-CVPR-2022} contains a semantic encoder that learns a descriptive latent space. The output of this encoder is used to condition an instance of DDIM. The proposed method allows DDPMs to perform well on tasks such as interpolation or attribute manipulation. %For both tasks the experiments are conducted on CelebA-HQ.

% 47. Come-Closer-Diffuse-Faster: Accelerating Conditional Diffusion Models for Inverse Problems through Stochastic Contraction 
Chung \etal~\cite{chung-CVPR-2022} introduce an algorithm for sampling, which reduces the number of steps required for the conditional case. Compared to the standard case, where the reverse process starts from Gaussian noise, their approach first executes one forward step to obtain an intermediary noised image and resumes the sampling from this point on. The approach is tested on inpainting, super-resolution, and magnetic resonance imaging (MRI) reconstruction. % in both cases FFHQ and AFHQ are used. Further, the authors evaluate the method on MRI reconstruction using the fastMRI knee data set.

% 49.DiffusionCLIP: Text-Guided Diffusion Models for Robust Image Manipulation
In \cite{kim-CVPR-2022}, the authors fine-tune a pretrained DDIM to generate images according to a text description. They propose a local directional CLIP loss that basically enforces the direction between the generated image and the original image to be as close as possible to the direction between the reference (original domain) and target text (target domain). The tasks considered in the evaluation are image translation between unseen domains, and multi-attribute transfer. % For these experiments the data sets are the following: ImageNet, CelebA-HQ, AFHQ-Dog, LSUN-Bedroom, LSUN- Church.

% 70class-cond+image2image\textbf{Understanding DDPM Latent Codes Through Optimal Transport}. 
Starting from the formulation of diffusion models as SDEs of Meng \etal~\cite{meng-arXiv-2021}, Khrulkov \etal~\cite{khrulkov-arXiv-2022} investigate the latent space and the resulting encoder maps. As per Monge formulation, it is shown that these encoder maps are the optimal transport maps, but this is demonstrated only for multivariate normal distributions. The authors further support this with numerical experiments, as well as practical experiments, using the model implementation of Dhariwal \etal~\cite{dhariwal-NeurIPS-2021}. 

% 72super-resolution+inpainting\textbf{Conditional Simulation Using Diffusion Schrödinger Bridges} 
Shi \etal~\cite{shi-arXiv-2022a} start by observing how an unconditional score-based diffusion model can be formulated as a Schr\"{o}dinger bridge, which can be solved using a modified version of Iterative Proportional Fitting. The previous method is reformulated to accept a condition, thus making conditional synthesis possible. Further adjustments are made to the iterative algorithm in order to optimize the time required to converge. The method is first validated with synthetic data from Kovachki \etal~\cite{kovachki-arXiv-2020}, showing improved capabilities in estimating the ground truth. The authors also conduct experiments on super-resolution, inpainting, and biochemical oxygen demand, the latter task being inspired by Marzouk \etal~\cite{marzouk-arXiv-2016}.
% they conduct experiments on biochemical oxygen demand (BOD), inspired from Marzouk \etal~\cite{marzouk-arXiv-2016}, producing more accurate results. Finally, image generation is evaluated on super resolution and inpainting on both MNIST and CelebA. 

% 74uncond+class-cond+text-cond\textbf{Retrieval-Augmented Diffusion Models}. 
Inspired by the Retrieval Transformer \cite{borgeaud-arXiv-2021}, Blattmann \etal~\cite{blattmann-arXiv-2022} propose a new method for training diffusion models. First, a set of similar images is fetched from a database using a nearest neighbor algorithm. The images are further encoded by an encoder with fixed parameters and projected into the CLIP \cite{radford-ICML-2021} feature space. Finally, the reverse process of the diffusion model is conditioned on this latent space. The method can be further extended to use other conditional signals, \eg~text, by simply enhancing the latent space with the encoded representation of the signal. %Comparing the performance with others SOTA diffusion models, as well as GAN based method, on ImageNet, it obtains better FID and IS scores than most of them, having competitive results at least. 

% 78uncond+class-cond+editing\textbf{Accelerating Diffusion Models via Early Stop of the Diffusion Process} 
Lyu \etal~\cite{lyu-arXiv-2022} introduce a new technique to reduce the number of sampling steps of diffusion models, boosting the performance at the same time. The idea is to stop the diffusion process at an earlier step. As the sampling cannot start from a random Gaussian noise, a GAN or VAE model is used to encode the last diffused image into a Gaussian latent space. The result is then decoded into an image which can be diffused into the starting point of the backward process. % The experiments are carried out on multiple data sets (CIFAR-10, CelebA, ImageNet, LSUN-Bedroom and LSUN-Cat), at various resolutions, showing great results in fewer steps on single-class data sets, while for multi-class data sets it requires more denoising steps. 

% 88class-cond+segmentation\textbf{Diffusion Models as Plug-and-Play Priors}. 
The aim of Graikos \etal~\cite{graikos-arXiv-2022} is to separate diffusion models into two independent parts, a prior (the base part) and a constraint (the condition). This enables models to be applied on various tasks without further training. Changing the equation of DDPMs from \cite{ho-NeurIPS-2020} leads to independently training the model and using it in a conditional setting, given that the constraint becomes differentiable. The authors conduct experiments on conditional image synthesis and image segmentation.
% An experiment is carried out on image synthesis with a diffusion model trained on MNIST and conditioned for various constraints: thin/thick digits, specific class or vertical/horizontal symmetry. Other experiments settings are: conditioning for facial characteristics on FFHQ-256 data set with a ResNet18 classifier pretrained on CelebA, image segmentation using EnviroAtlas data set as well as solving the traveling salesman problem.

\subsection{Medical Image Generation and Translation}
%2021
% 89\textbf{Diffusion Models for Implicit Image Segmentation Ensembles} 
Wolleb \etal~\cite{wolleb-arXiv-2021} introduce a method based on diffusion models for image segmentation within the context of brain tumor segmentation. The training consists of diffusing the segmentation map, then denoising it to obtain the original image. During the backward process, the brain MR image is concatenated into the intermediate denoising steps in order to be passed through the U-Net model, thus conditioning the denoising process on it. Furthermore, for each input, the authors propose to generate multiple samples, which will be different due to stochasticity. Thus, the ensemble can generate a mean segmentation map and its variance (associated with the uncertainty of the map). %Evaluation is done on BRATS2020 data set using the DDPM from \cite{nichol-ICML-2021}, results being comparable to nnU-Net and SegNet baselines. 

% 108\textbf{SOLVING INVERSE PROBLEMS IN MEDICAL IMAGING WITH SCORE-BASED GENERATIVE MODELS} 
Song \etal~\cite{song-arXiv-2021} introduce a method for score-based models that is able to solve inverse problems in medical imaging, \ie~reconstructing images from measurements. First, an unconditional score model is trained. Then, a stochastic process of the measurement is derived, which can be used to infuse conditional information into the model via a proximal optimization step. Finally, the matrix that maps the signal to the measurement is decomposed to allow sampling in closed-form. The authors carry out multiple experiments on different medical image types, including computed tomography (CT), low-dose CT and MRI.
%: CT (on Lung Image Database Consortium and Low Dose CT Image and Projection data set), MAR (on synthetic data from \cite{yu-ITMI-2020}) and  MRI (on Brain Tumor Segmentation), leading to competitive results. 

% 109\textbf{Score-based diffusion models for accelerated MRI}. 
Within the area of medical imaging, but focusing on reconstructing the images from accelerated MRI scans, Chung \etal~\cite{chung-MIA-2022} propose to solve the inverse problem using a score-based diffusion model. A score model is pretrained only on magnitude images in an unconditional setting. Then, a variance exploding SDE solver \cite{song-ICLR-2021} is employed in the sampling process. By adopting a Predictor-Corrector algorithm \cite{song-ICLR-2021} interleaved with a data consistency mapping, the split image (real and imaginary parts) is fed through, enabling conditioning the model on the measurement. Furthermore, the authors present an extension of the method which enables conditioning on multiple coil-varying measurements. % In order to demonstrate the effectiveness of the method, it was compared to a wide range of baselines: total variation reconstruction \cite{block-MRM-2007}, U-Net based model trained in supervised setting \cite{zbontar-arXiv-2018}, DuDoRNet \cite{zhou-CVPR-2020} and E2Evarnet \cite{sriram-MICCAI-2020}. The experiments were done on fastMRI knee data set. 

%2022
% 93\textbf{Unsupervised Medical Image Translation with Adversarial Diffusion Models}. 
\"{O}zbey \etal~\cite{ozbey-arXiv-2022} propose a diffusion model with adversarial inference. In order to increase each diffusion step, and thus make fewer steps, inspired by \cite{xiao-arXiv-2021}, the authors employ a GAN model in the reverse process to estimate the denoised image at every step. Using a similar method as \cite{zhu-ICCV-2017}, they introduce a cycle-consistent architecture to allow training on unpaired data sets. %, but are coupled with two further diffusion-based generators. 
%While the diffusive models are composed of U-Nets, the non-diffusive ones use a ResNet architecture. The model is evaluated for image generation on multi-contrast brain MRI data (IXI from \url{https://brain-development.org/ixi-data set}) and a multi-modal pelvic MRI-CT data set \cite{nyholm-IOMP-2018}, outperforming state-of-the-art models. 

% 110\textbf{Unsupervised Denoising of Retinal OCT with Diffusion Probabilistic Model}. 
The aim of Hu \etal~\cite{hu-SPIE-2022} is to remove the speckle noise in optical coherence tomography (OCT) b-scans. The first stage is represented by a method called self-fusion, as described in \cite{oguz-SPIE-2020}, where additional b-scans close to the given 2D slice of the input OCT volume are selected. The second stage consists of a diffusion model whose starting point is the weighted average of the original b-scan and its neighbors. Thus, the noise can be removed by sampling a clean scan. %The training data consists of 6 optic nerve head volumes of the human retina, while for testing 6 fovea volumes were used, each with 500 b-scans. The baseline used for comparison is PMFN by Hu \etal~\cite{hu-OMIA-2020}, results showing that the proposed method is superior on all metrics (SNR, CNR and ENL). 

\subsection{Anomaly Detection in Medical Images}

Auto-encoders are widely used for anomaly detection \cite{Ionescu-CVPR-2019}. Since diffusion models can be seen as a particular type of VAEs, it seems natural to employ diffusion models for the same tasks as VAEs. So far, diffusion models have shown promising results in detecting anomalies in medical images, as further discussed below.

Wyatt \etal~\cite{Wyatt-CVPRW-2022} train a DDPM on healthy medical images. The anomalies are detected at inference time by subtracting the generated image from the original image. The work also proves that using simplex noise instead of Gaussian noise yields better results for this type of task.

% 97\textbf{Diffusion Models for Medical Anomaly Detection}. 
Wolleb \etal~\cite{wolleb-arXiv-2022a} propose a weakly-supervised method based on diffusion models for anomaly detection in medical images. Given two unpaired images, one healthy and one with lesions, the former is diffused by the model. Then, the denoising process is guided by the gradient of a binary classifier in order to generate the healthy image. Finally, the sampled healthy image and the one containing lesions are subtracted to obtain the anomaly map. %Using the same configuration of the diffusion model as \cite{dhariwal-NeurIPS-2021}, experiments were carried out on CheXpert and BRATS2020 data sets. Compared to Fixed-Point GAN \cite{siddiquee-ICCV-2019} and a VAE-based model \cite{zimmerer-arXiv-2018}, the method shows improved results, images having a better quality i.e. more realistic. 

% 111\textbf{Fast Unsupervised Brain Anomaly Detection and Segmentation with Diffusion Models}
Pinaya \etal~\cite{pinaya-arXiv-2022} propose a diffusion-based method for detecting anomalies in brain scans, as well as segmenting those regions. The images are encoded by a VQ-VAE \cite{van-NIPS-2017}, and the quantized latent representation is obtained from a codebook. The diffusion model operates in this latent space. Averaging the intermediate samples from median steps of the backward process and then applying a precomputed threshold map, a binary mask implying the anomaly location is created. Starting the backward process from the middle, the binary mask is used to denoise the anomalous regions, while maintaining the rest. Finally, the sample at the final step is decoded, resulting in a healthy image. The segmentation map of the anomaly is obtained by subtracting the input image and the synthesized image. %Two experiments were done to assess the performance of the method: one on synthetic data from MedNIST data set, and the other one using UK Biobank \cite{sudlow-PLOSM-2015} images for training and testing on WMH \cite{kuijf-ITMI-2019}, BRATS \cite{bakas-arXiv-2018}, MSLUB \cite{lesjak-Neuroinformatics-2018} and white matter hyperintensities from UK Biobank. Reporting the DICE-score, the method has a competitive performance while being significantly more time-efficient.

% Paper from Pedro (Uni of Edin)
Sanchez \etal~\cite{Sanchez-DGM4Miccai-2022} follow the same principle for detecting and segmenting anomalies in medical images: a diffusion model generates healthy samples which are then subtracted from the original images. The input image is diffused using the model, reversing the denoising equation and nullifying the condition, and then the backward conditioned process is applied. Utilizing a classifier-free model, the guidance is achieved through an attention mechanism integrated in the U-Net. The training utilizes both healthy and unhealthy examples.

\subsection{Video Generation}
%video
The recent progress towards making diffusion models more efficient has enabled their application in the video domain. We next present works applying diffusion models to video generation.

Ho \etal~\cite{ho-ICLRW-2022} introduce diffusion models to the task of video generation. When compared to the 2D case, the changes are applied only to the architecture. The authors adopt the 3D U-Net from \cite{cicek-MICCAI-2016}, presenting results in unconditional and text conditional video generation. Longer videos are generated in an autoregressive manner, where the latter video chunks are conditioned on the previous ones.

Yang \etal~\cite{yang-arXiv-2022} generate videos frame by frame, using diffusion models. The reverse process is entirely conditioned on a context vector provided by a convolutional recurrent neural network. The authors perform an ablation study to decide if predicting the residual of the next frame returns better results than the case of predicting the actual frame. The conclusion is that the former option works better. 

H{\"o}ppe \etal~\cite{hoppe-arXiv-2022} present random mask video diffusion (RaMViD), a method which can be used for video generation and infilling. The main contribution of their work is a novel strategy for training, which randomly splits the frames into masked and unmasked frames. The unmasked frames are used to condition the diffusion, while the masked ones are diffused by the forward process. 

The work of Harvey \etal~\cite{harvey-arXiv-2022} introduces flexible diffusion models, a type of diffusion model that can be used with multiple sampling schemes for long video generation. As in \cite{hoppe-arXiv-2022}, the authors train a diffusion model by randomly choosing the frames used in the diffusion and those used for conditioning the process. After training the model, they investigate the effectiveness of multiple sampling schemes, concluding that the sampling choice depends on the data set.

\subsection{Other Tasks}

There are some pioneering works applying diffusion models to new tasks, which have been scarcely explored via diffusion modeling. We gather and discuss such contributions below.

% 8. Diffusion Probabilistic Models for 3D Point Cloud Generation. 
Luo \etal~\cite{luo-CVPR-2021} apply diffusion models on 3D point cloud generation, auto-encoding, and unsupervised representation learning. They derive an objective function from the variational lower bound of the likelihood of point clouds conditioned on a shape latent. The experiments are conducted using PointNet \cite{charles-CVPR-2017} as the underlying architecture.

% 18. 3D Shape Generation and Completion through Point-Voxel Diffusion
Zhou \etal~\cite{zhou-ICCV-2021} introduce Point-Voxel Diffusion (PVD), a novel method for shape generation which applies diffusion on point-voxel representations. The approach addresses the tasks of shape generation and completion on the ShapeNet and PartNet data sets.

% 28. Score-Based Generative Classifiers
Zimmermann \etal~\cite{zimmermann-arXiv-2021} show a strategy to apply score-based models for classification. They add the image label as a conditioning variable to the score function and, thanks to the ODE formulation, the conditional likelihood can be computed at inference time. Thus, the prediction is the label with the maximum likelihood. Further, they study the impact of this type of classifier on out-of-distribution scenarios considering common image corruptions and adversarial perturbations. %The results are reported on CIFAR-10 using the same architecture as in \cite{song-NeurIPS-2021}.

%2021
% 100\textbf{DiffuseMorph: Unsupervised Deformable Image Registration Along Continuous Trajectory Using Diffusion Models}
Kim \etal~\cite{kim-arXiv-2021} propose to solve the image registration task using diffusion models. This is achieved via two networks, a diffusion network, as per \cite{ho-NeurIPS-2020}, and a deformation network that is based on U-Net, as described in \cite{balakrishnan-CVPR-2018}. Given two images (one static, one moving), the role of the former network is to assess the deformation between the two images, and feed the result to the latter network, which predicts the deformation fields, enabling sample generation. This method also has the ability to synthesize the deformations through the whole transition. The authors carried out experiments for different tasks, one on 2D facial expressions and one on 3D brain images. The results confirm that the model is capable of producing qualitative and accurate registration fields.

%2022
% 40.  Diffusion Models for Counterfactual Explanations
Jeanneret \etal~\cite{jeanneret-arXiv-2022} apply diffusion models for counterfactual explanations. The method starts from a noised query image, and generates a sample with an unconditional DDPM. With the generated sample, the gradients required for guidance are computed. Then, one step of the reversed guided process is applied. The output is further used in the next reverse steps. %The authors run experiments on CelebA, using DDPM.

% 113. Diffusion Causal Models for Counterfactual Estimation
Sanchez \etal~\cite{sanchez-CLeaR-2022} adapt the work of Dhariwal \etal~\cite{dhariwal-NeurIPS-2021} for counterfactual image generation. As in \cite{dhariwal-NeurIPS-2021}, the denoising process is guided by classifier gradients to generate samples from the desired counterfactual class. The key contribution is the algorithm used to retrieve the latent representation of the original image, from where the denoising process starts. Their algorithm inverts the deterministic sampling procedure of \cite{song-ICLR-2021b} and maps each original image to a unique latent representation.

% 62\textbf{Diffusion Models for Adversarial Purification}. 
Nie \etal~\cite{nie-arXiv-2022} demonstrate how a diffusion model can be used as a defensive mechanism for adversarial attacks. Given an adversarial image, it gets diffused up until an optimally computed time step. The result is then reversed by the model, producing a purified sample at the end. To optimize the computations of solving the reverse-time SDE, the adjoint sensitivity method of Li \etal~\cite{li-AISTATS-2020} is used for the gradient score calculations. %Carrying out the experiments using three data sets (CIFAR-10, ImageNet and CelebA-HQ) with three classifiers (ResNet, WideResNet and ViT) on the benchmark RobustBench, they report strong results, noticeably outperforming previous methods. 

% 80\textbf{Few-Shot Diffusion Models}. 
In the context of few-shot learning, an image generator based on diffusion models is proposed by Giannone \etal~\cite{giannone-arXiv-2022}. Given a small set of images that condition the synthesis, a visual transformer encodes these, and the resulting context representation is integrated (via two different techniques) into the U-Net model employed in the denoising process. % Experiments were carried out with probabilistic diffusion models as per (\cite{ho-NeurIPS-2020}, \cite{nichol-ICML-2021}), comparing with both conditional and unconditional baselines on Omniglot, FS-CIFAR100 \cite{khrulkov-arXiv-2022}, miniImageNet and CelebA. It produced great results, considerably outperforming on generating images of unseen classes. 

% 87In \textbf{Semantic Image Synthesis via Diffusion Models}
Wang \etal~\cite{wang-arXiv-2022b} present a framework based on diffusion models for semantic image synthesis. Leveraging the U-Net architecture of diffusion models, the input noise is supplied to the encoder, while the semantic label map is passed to the decoder using multi-layer spatially-adaptive normalization operators \cite{Park-CVPR-2019}. To further improve the sampling quality and the condition on the semantic label map, an empty map is also supplied to the sampling method to generate the unconditional noise. Then, the final noise uses both estimates. % Carrying out the experiments on Cityscapes, ADE20K and CelebAMask-HQ benchmark data sets, the method outperforms GAN-based models. 

% 91\textbf{Restoring Vision in Adverse Weather Conditions with Patch-Based Denoising Diffusion Models}. 
Concerning the task of restoring images negatively affected by various weather conditions (\eg~snow, rain), {\"O}zdenizci \etal~\cite{ozdenizci-arXiv-2022} demonstrate how diffusion models can be used. They condition the denoising process on the degraded image by concatenating it channel-wise to the denoised sample, at every time step. In order to deal with varying image sizes, at every step, the sample is divided into overlapping patches, passed in parallel through the model, and combined back by averaging the overlapping pixels. The employed diffusion model is based on the U-Net architecture, as presented in \cite{ho-NeurIPS-2020,song-ICLR-2021}, but modified to accept two concatenated images as input. % The evaluation was done on Snow100K, Outdoor-Rain and RainDrop benchmark data sets, showing that the method outperforms previous state-of-the-art models, as well as demonstrating strong performance in real-world scenarios. 

% 99\textbf{Denoising Diffusion Restoration Models}. 
Formulating the task of image restoration as a linear inverse problem, Kawar \etal~\cite{kawar-arXiv-2022} propose the use of diffusion models. Inspired by Kawar \etal~\cite{kawar-NeurIPS-2021}, the linear degradation matrix is decomposed using singular value decomposition, such that both the input and the output can be mapped onto the spectral space of the matrix where the diffusion process is carried out. Leveraging the pretrained diffusion models from \cite{ho-NeurIPS-2020} and \cite{dhariwal-NeurIPS-2021}, the evaluation is conducted on various tasks: super-resolution, deblurring, colorization and inpainting. %) showed that it outperformed all the baselines (RED \cite{romano-SIAM-2017}, DGP \cite{pan-TPAMI-2021} and SNIPS \cite{kawar-NeurIPS-2021}) for few network evaluations.

\subsection{Theoretical Contributions}

%16. A Variational Perspective on Diffusion-Based Generative Models and Score Matching
Huang \etal~\cite{huang-NeurIPS-2021} demonstrate how the method proposed by Song \etal~\cite{song-ICLR-2021} is linked with maximizing a lower bound on the marginal likelihood of the reverse SDE. Moreover, they verify their theoretical contribution with image generation experiments on CIFAR-10 and MNIST.

\section{Closing Remarks and Future Directions}
\label{sec_experiments}
In this paper, we reviewed the advancements made by the research community in developing and applying diffusion models to various computer vision tasks. We identified three primary formulations of diffusion modeling based on: DDPMs, NCSNs, and SDEs. Each formulation obtains remarkable results in image generation, surpassing GANs while increasing the diversity of the generated samples. The outstanding results of diffusion models are achieved while the research is still in its early phase. Although we observed that the main focus is on conditional and unconditional image generation, there are still many tasks to be explored and further improvements to be realized.

\noindent {\bf Limitations.} The most significant disadvantage of diffusion models remains the need to perform multiple steps at inference time to generate only one sample. Despite the important amount of research conducted in this direction, GANs are still faster at producing images. Other issues of diffusion models can be linked to the commonly used strategy to employ CLIP embeddings for text-to-image generation. For example, Ramesh \etal~\cite{ramesh-arXiv-2022} highlight that their model struggles to generate readable text in an image and motivate the behavior by stating that CLIP embeddings do not contain information about spelling. Therefore, when such embeddings are used for conditioning the denoising process, the model can inherit this kind of issues.

\noindent {\bf Future directions.} To reduce the uncertainty level, diffusion models generally avoid taking large steps during sampling. Indeed, taking small steps ensures the data sample generated at each step is explained by the learned Gaussian distribution. A similar behavior is observed when applying gradient descent to optimize neural networks. Indeed, taking a large step in the negative direction of the gradient, \ie~using a very large learning rate, can lead to updating the model to a region with high uncertainty, having no control over the loss value. In future work, transferring update rules borrowed from efficient optimizers to diffusion models could perhaps lead to a more efficient sampling (generation) process.

Aside from the current tendency of researching more efficient diffusion models, future work can study diffusion models applied in other computer vision tasks, such as image dehazing, video anomaly detection, or visual question answering. Even if we found some works studying anomaly detection in medical images \cite{Wyatt-CVPRW-2022,wolleb-arXiv-2022a,pinaya-arXiv-2022}, this task could also be explored in other domains, such as video surveillance or industrial inspection.

An interesting research direction is to assess the quality and utility of the representation space learned by diffusion models in discriminative tasks. This could be carried out in at least two distinct ways. In a direct way, by learning some discriminative model on top of the latent representations provided by a denoising model, to address some classification or regression task. In an indirect way, by augmenting training sets with realistic samples generated by diffusion models. The latter direction might be more suitable for tasks such as object detection, where inpainting diffusion models could do a good job at blending in new objects in images.

Another future work direction is to employ conditional diffusion models to simulate possible futures in video. The generated videos could further be given as input to reinforcement learning models.

Recent diffusion models \cite{ho-ICLRW-2022} have shown impressive text-to-video synthesis capabilities compared to the previous state of the art, significantly reducing the number of artifacts and reaching an unprecedented generative performance. However, we believe this direction requires more attention in future work, as the generated videos are rather short. Hence, modeling long-term temporal relations and interactions between objects remains an open challenge.

In future, the research on diffusion models can also be expanded towards learning multi-purpose models that solve multiple tasks at once. Creating a diffusion model to generate multiple types of outputs, while being conditioned on various types of data, \eg~text, class labels or images, might take us closer to understanding the necessary steps towards developing artificial general intelligence (AGI).

%\noindent{\bf Conclusion.} This work depicts the latest developments concerning diffusion models. We describe the three most common formulations of the model and the connections with the other generative models. After that, we group according to the task the methods published so far, and we describe each paper in a few sentences, summarizing their contribution.

\ifCLASSOPTIONcompsoc
  \section*{Acknowledgments}
\else
  \section*{Acknowledgment}
\fi

This work was supported by a grant of the Romanian Ministry of Education and Research, CNCS - UEFISCDI, project no. PN-III-P2-2.1-PED-2021-0195, contract no. 690/2022, within PNCDI III.

%%%%%%%%% REFERENCES
{\small
\bibliographystyle{ieeetr}
\bibliography{references}
}

\begin{IEEEbiography}[{\includegraphics[width=0.6in,height=0.78in,clip,keepaspectratio]{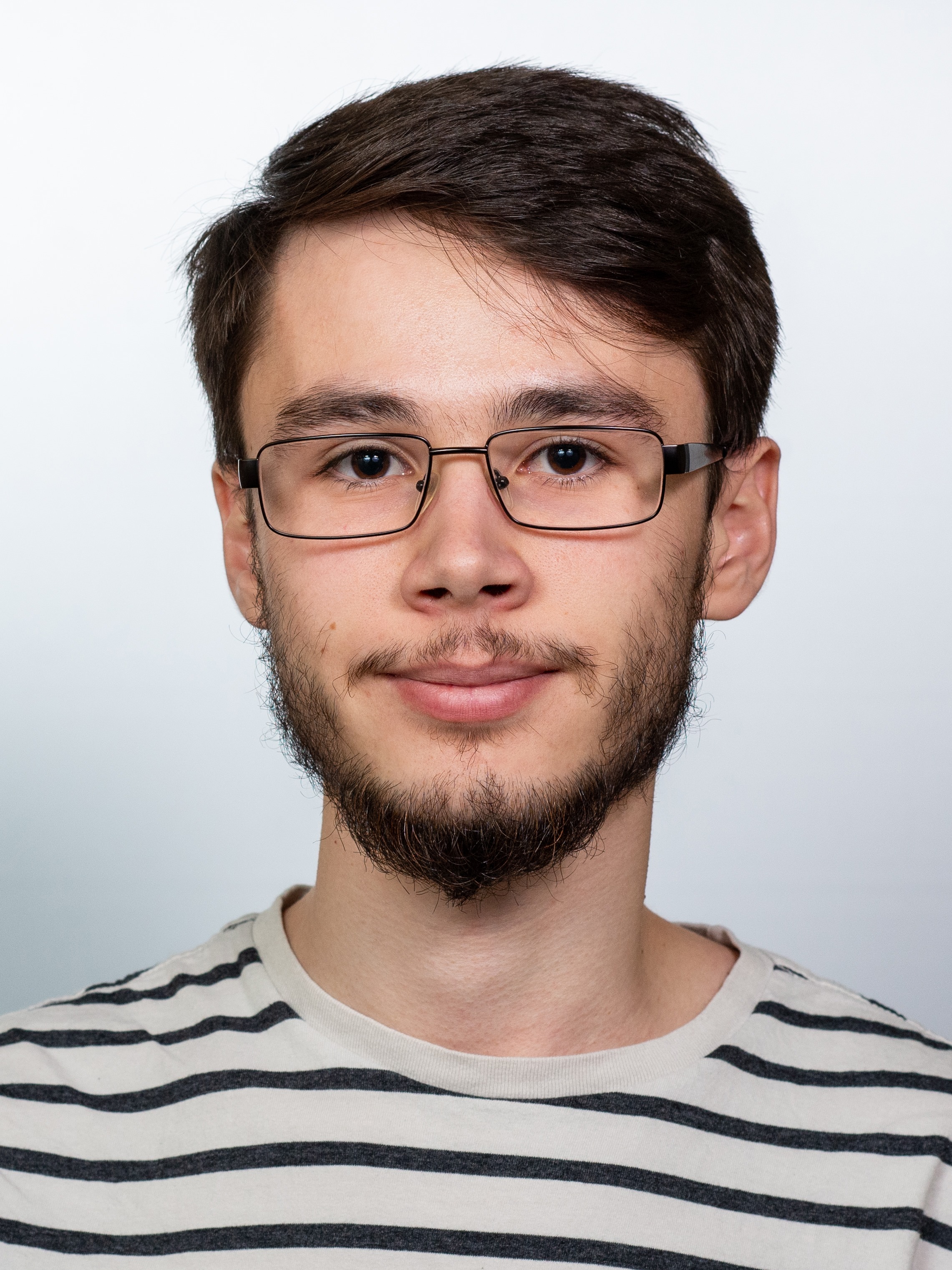}}]
{Florinel-Alin Croitoru} is a Ph.D. student at the University of Bucharest, Romania. He obtained his bachelor's degree from the Faculty of Mathematics and Computer Science of the University of Bucharest in 2019. In 2021, he obtained his masters degree in Artificial Intelligence with a thesis on action spotting in football videos. His domains of interest include machine learning, computer vision and deep learning.
\end{IEEEbiography}

\begin{IEEEbiography}[{\includegraphics[width=0.6in,height=0.78in,clip,keepaspectratio]{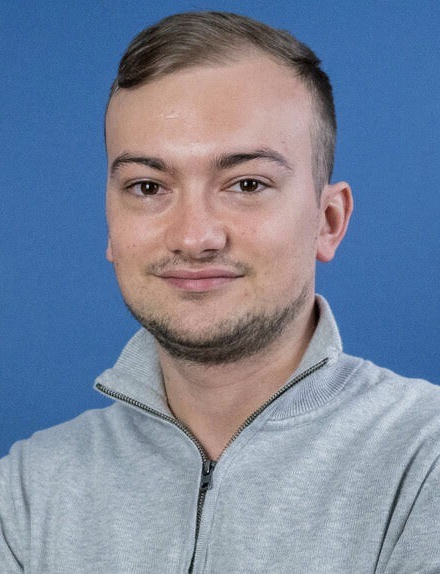}}]
{Vlad Hondru} is a Ph.D. student at the University of Bucharest, Romania. He obtained his bachelor's degree from the University of Manchester in Mechatronic Engineering, then he graduated from Imperial College London, studying towards an MSc in Computing Science, with a Visual Computing and Robotics specialization, focusing on Artificial Intelligence. He did a year-long placement at Rolls-Royce as a software engineer, as well as undertaking a summer internship within the Robotics Group of the University of Manchester. He currently works as a machine learning engineer, developing NLP products.
\end{IEEEbiography}

%\vspace{-1cm}
\begin{IEEEbiography}[{\includegraphics[width=0.6in,height=0.78in,clip,keepaspectratio]{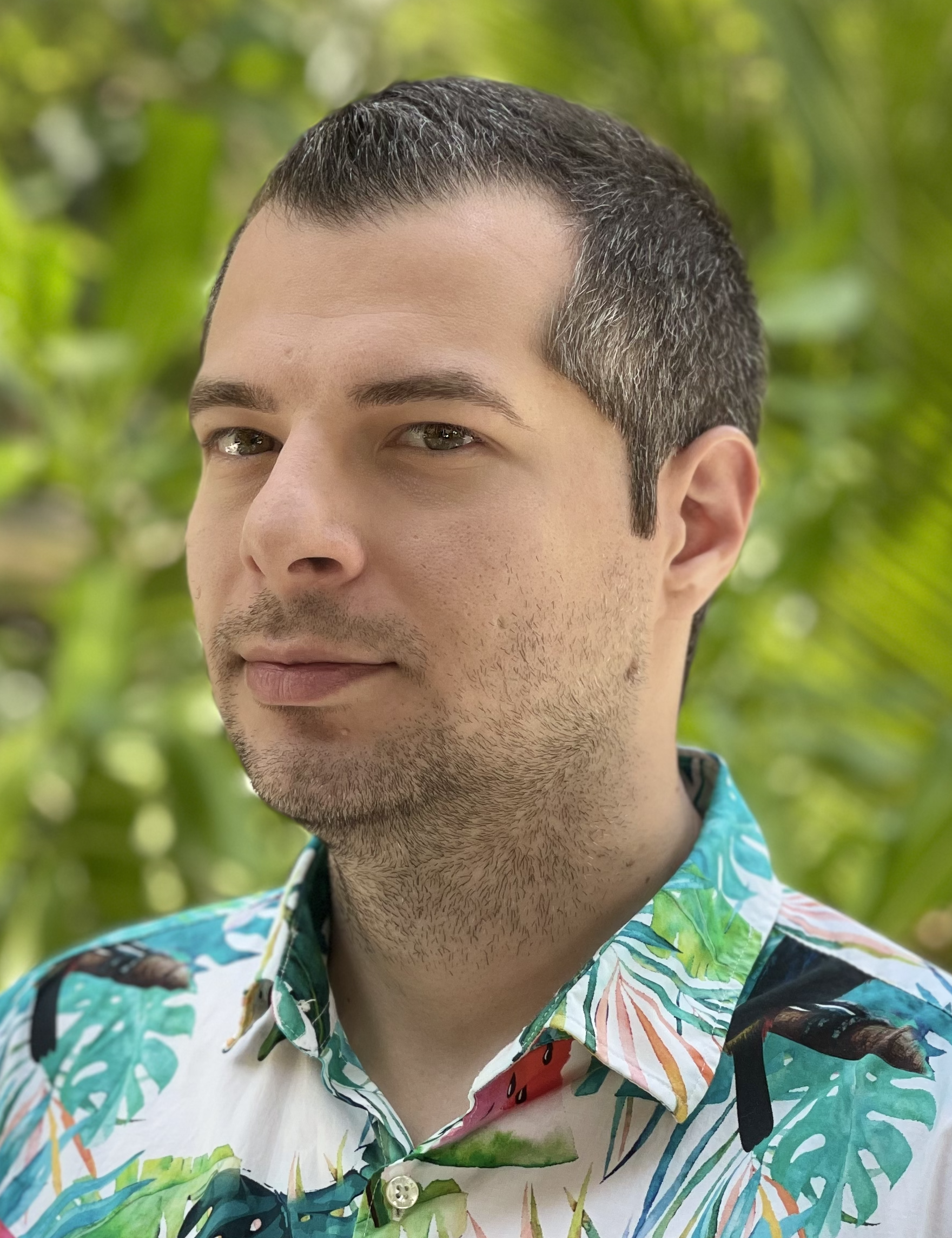}}]
{Radu Ionescu} is professor at the University of Bucharest, Romania. He completed his Ph.D.~at the University of Bucharest in 2013, receiving the 2014 Award for Outstanding Doctoral Research from the Romanian Ad Astra Association.
His research interests include machine learning, computer vision, image processing, computational linguistics and medical imaging. He published over 100 articles at international venues (including CVPR, NeurIPS, ICCV, ACL, EMNLP, NAACL, TPAMI, IJCV, CVIU), and a research monograph with Springer. Radu received the ``Caianiello Best Young Paper Award'' at ICIAP 2013. % for the paper entitled ``Kernels for Visual Words Histograms''. 
Radu also received the 2017 ``Young Researchers in Science and Engineering'' Prize for young Romanian researchers and the ``Danubius Young Scientist Award 2018 for Romania''. %He participated at several international competitions obtaining top ranks: 4th place in the Facial Expression Recognition Challenge of WREPL 2013, 3rd place in the Native Language Identification Shared Task of BEA-8 2013, 2nd place in the Arabic Dialect Identification Shared Task of VarDial 2016, 
%1st place in the Arabic Dialect Identification Shared Tasks of VarDial 2017 and 2018, 1st place in the Native Language Identification Shared Task of BEA-12 2017.
\end{IEEEbiography}

%\vspace{-1cm}
\begin{IEEEbiography}[{\includegraphics[width=0.6in,height=0.78in,clip,keepaspectratio]{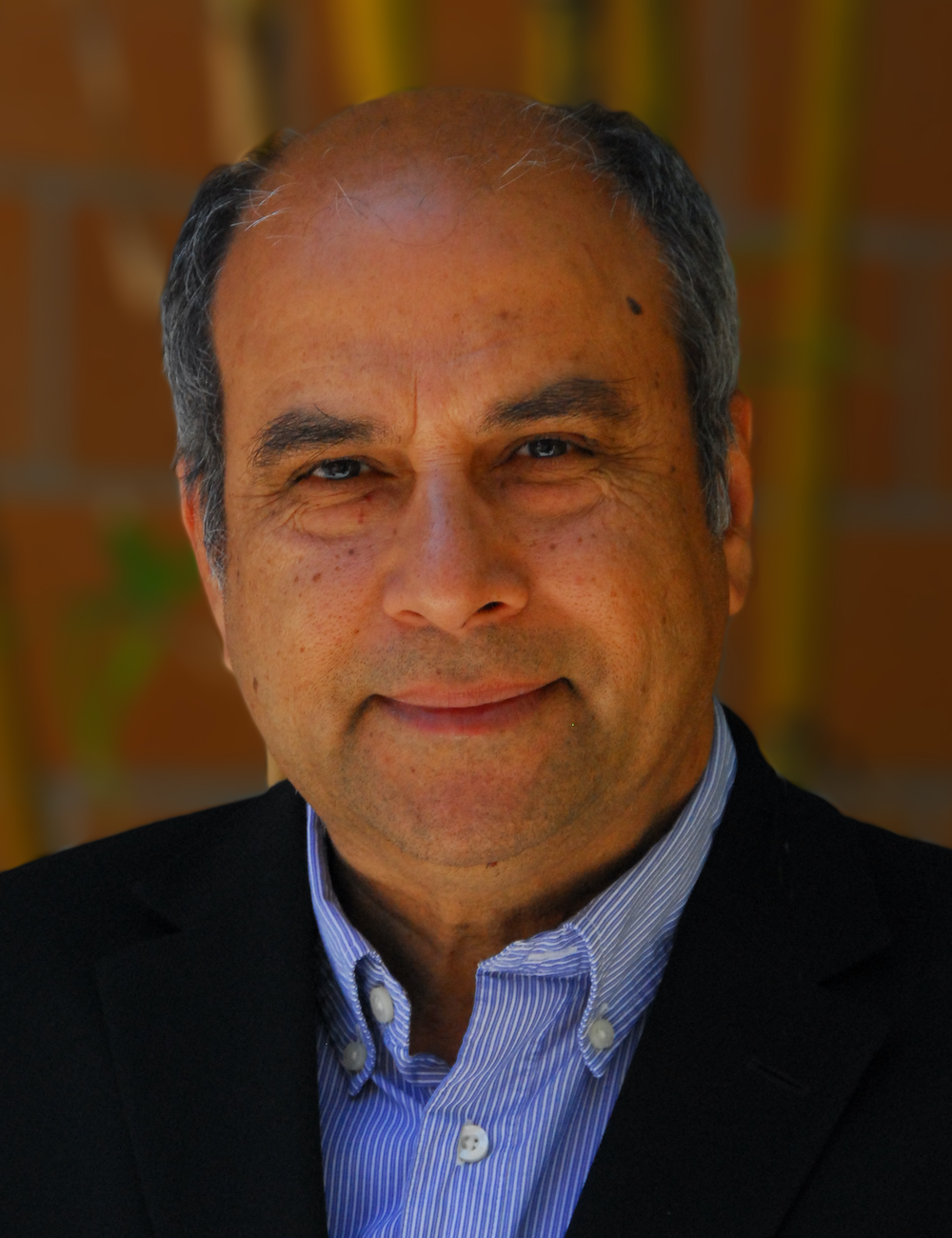}}]
{Mubarak Shah} is the UCF Trustee chair professor and the founding director of the Center for Research in Computer Vision at the University of Central Florida (UCF). He is a fellow of the NAI, IEEE, AAAS, IAPR and SPIE. He is an editor of an international book series on video computing, was editor-in-chief of Machine Vision and Applications and an associate editor of ACM Computing Surveys and IEEE TPAMI. %He was the program cochair of CVPR 2008, an associate editor of the IEEE TPAMI and a guest editor of the special issue of the International Journal of Computer Vision on Video Computing. 
His research interests include video surveillance, visual tracking, human activity recognition, visual analysis of crowded scenes, video registration, UAV video analysis, among others. He has served as an ACM distinguished speaker and IEEE distinguished visitor speaker. He is a recipient of ACM SIGMM Technical Achievement award; IEEE Outstanding Engineering Educator Award; Harris Corporation Engineering Achievement Award; an honorable mention for the ICCV 2005 ``Where Am I?'' Challenge Problem; 2013 NGA Best Research Poster Presentation; 2nd place in Grand Challenge at the ACM Multimedia 2013 conference; and runner up for the best paper award in ACM Multimedia Conference in 2005 and 2010. At UCF, he has received the Pegasus Professor Award, University Distinguished Research Award, Faculty Excellence in Mentoring Doctoral Students, Scholarship of Teaching and Learning Award, Teaching Incentive Program Award, Research Incentive Award.
\end{IEEEbiography}

\begin{appendices}

\section{Variational bound.}
\label{appendix_vlb}

We emphasize that the derivation presented below is also shown in \cite{sohl-icml-2015, ho-NeurIPS-2020}. The variational bound of the log density of the data can be derived as in the case of VAEs \cite{Kingma-ICLR-2014b}, where the latent variables are the noisy images $x_{1:T}$ and the observed variable is the original image $x_0$.
We start by writing the log likelihood of the data $\log{p_\theta(x_0)}$ as the log marginal of the joint probability $p_\theta(x_{0:T})$:
\begin{equation}\label{eq_marginal_prob}
\begin{split}
    \log{p_\theta(x_0)}\!&=\!\log{\int\!p_\theta(x_{0:T}) \partial x_{1:T}} \\
    &= \log{\int\!p_\theta(x_{0:T}) \cdot \frac{p(x_{1:T}|x_0)}{p(x_{1:T}|x_0)} \partial x_{1:T}} \\
    &= \log{\int p(x_{1:T}) \cdot \frac{p_\theta(x_{0:T})}{p(x_{1:T}|x_0)} \partial x_{1:T}} \\
    &= \log{\mathbb{E}_{x_{1:T} \sim p(x_{1:T}|x_0)}}\left[\frac{p_\theta(x_{0:T})}{p(x_{1:T}|x_0)}\right].
\end{split}
\end{equation}

Jensen's inequality states that, given a random variable $Y$ and a convex function $f$, the following is true:
\begin{equation}\label{jensen_ineq}
    f(\mathbb{E}[Y]) \leq \mathbb{E}[f(Y)].
\end{equation}

If we apply Eq.~\eqref{jensen_ineq} to Eq.~\eqref{eq_marginal_prob} and change the inequality sign because the $\log$ function is concave, then we obtain the following:
\begin{equation}\label{eq_negative_log_likelihood}
\begin{split}
    \log{p_\theta(x_0)} &\geq \mathbb{E}_{x_{1:T} \sim p(x_{1:T}|x_0)}\left[\log{\frac{p_\theta(x_{0:T})}{p(x_{1:T}|x_0)}}\right] \bigg\rvert \cdot(-1) \\
    -\log{p_\theta(x_0)} &\leq \mathbb{E}_{x_{1:T} \sim p(x_{1:T}|x_0)}\left[\log{\frac{p(x_{1:T}|x_0)}{p_\theta(x_{0:T})}}\right].
\end{split}
\end{equation}
Eq.~\eqref{eq_negative_log_likelihood} shows that we can minimize the right-hand side of the inequality, instead of minimizing the expected negative log likelihood of the data for our generative model. We focus further on this term such that, at the end, we will derive the objective from Eq.~\eqref{eq_ddpm_3}.

By definition, the forward and reverse processes are Markovian. Based on this, we can rewrite the probabilities from Eq.~\eqref{eq_negative_log_likelihood} as follows:
\begin{equation}\label{eq_prob_markovian}
\begin{split}
   p(x_{1:T}|x_0) &\!=\!p(x_T|x_{1:T-1}, x_0) \cdot p(x_{1:T-1}|x_0) \\
   &\!=\!p(x_T|x_{T-1})\!\cdot\!p(x_{T-1}|x_{1:T-2}, x_0)\!\cdot\!p(x_{1:T-2}|x_0)\\
   &\!=\!\dots\!=\!\prod_{t=1}^{T}p(x_t|x_{t-1}),\\
   p_\theta(x_{0:T}) &= p_\theta(x_0|x_{1:T}) \cdot p_\theta(x_{1:T}) \\ 
   &= p_\theta(x_0|x_1) \cdot p_\theta(x_1|x_{2:T}) \cdot p_\theta(x_{2:T})\\
   &= \dots = p_\theta(x_T) \prod_{t=1}^T{p_\theta(x_{t-1}|x_t)}.
\end{split}
\end{equation}
We replace the probabilities in the right-hand side of Eq.~\eqref{eq_negative_log_likelihood} with the products from Eq.~\eqref{eq_prob_markovian} and apply the $\log$ property that transforms the products into sums:
\begin{equation}\label{eq_vlb}
\begin{split}
    &\mathbb{E}_{p}\left[\log{\frac{p(x_{1:T}|x_0)}{p_\theta(x_{0:T})}}\right] =\\
    &=\mathbb{E}_{x_{1:T} \sim p(x_{1:T}|x_0)} \left[-\log{p_\theta(x_T)} + \sum_{t=1}^T\log{\frac{p(x_t|x_{t-1})}{p_{\theta}(x_{t-1}|x_t)}} \right].
\end{split}
\end{equation}
The term $p(x_t|x_{t-1})$ can be transformed using the Bayes rule into $\frac{p(x_{t-1}|x_t) \cdot p(x_t)}{p(x_{t-1})}$, but the true posterior of the forward process $p(x_{t-1}|x_t)$ is intractable. However, if we additionally condition on the initial image $x_0$, the posterior becomes tractable. Moreover, we know that $p(x_t|x_{t-1},x_0)=p(x_t|x_{t-1})$ is true because the forward process is Markovian. Hence, if we apply the Bayes rule on $p(x_t|x_{t-1},x_0)$, we will obtain the additional conditioning on the true posterior:
\begin{equation}
\label{eq_fwd_posterior}
p(x_t|x_{t-1},x_0) = \frac{p(x_{t-1}|x_{t},x_0) \cdot p(x_t|x_0)}{p(x_{t-1}|x_0)}.
\end{equation}
If we apply the derivation from Eq.~\eqref{eq_fwd_posterior} to Eq.~\eqref{eq_vlb} for all $t \geq 2$, the result is as follows:
\begin{equation}
\begin{split}
\mathcal{L}_{\scriptsize{\mbox{\emph{vlb}}}} =&\; \mathbb{E}_p \left[-\log{p_\theta(x_T)} + \sum_{t=1}^T\log{\frac{p(x_t|x_{t-1})}{p_{\theta}(x_{t-1}|x_t)}}\right]\\
=&\;\mathbb{E}_p[-\log{p_\theta(x_T)}]+\mathbb{E}_p\left[\sum_{t=2}^{T}\log{\frac{p(x_{t-1}|x_t,x_0)}{p_\theta(x_{t-1}|x_t)}}\right]\\
&+\mathbb{E}_p\left[ \sum_{t=2}^{T}\log{\frac{p(x_t|x_0)}{p(x_{t-1}|x_0)}} + \log{\frac{p(x_1|x_0)}{p_\theta(x_0|x_1)}}\right].
\end{split}
\end{equation}
Observing that the terms of the second sum cancel out, $\mathcal{L}_{\scriptsize{\mbox{\emph{vlb}}}}$ becomes:
\begin{equation}
\begin{split}
\mathcal{L}_{\scriptsize{\mbox{\emph{vlb}}}} =&\; \mathbb{E}_p[-\log{p_\theta(x_T)}]
+ \mathbb{E}_p\left[\sum_{t=2}^{T}\log{\frac{p(x_{t-1}|x_t,x_0)}{p_\theta(x_{t-1}|x_t)}}\right]
\\&+\mathbb{E}_p\left[ \log{\frac{p(x_T|x_0)}{p(x_{1}|x_0)}} + \log{\frac{p(x_1|x_0)}{p_\theta(x_0|x_1)}}\right].
\end{split}
\end{equation}
Finally, if we rearrange the terms and transform the log rates into Kulback-Leibler divergences, the result is the formulation from Eq.~\eqref{eq_ddpm_3}:
\begin{equation}
\begin{split}
\mathcal{L}_{\scriptsize{\mbox{\emph{vlb}}}} =&\;\mathbb{E}_p\left[-\log{p_\theta(x_0|x_1)}\right] + \mathbb{E}_p\left[\sum_{t=2}^{T}\log{\frac{p(x_{t-1}|x_t,x_0)}{p_\theta(x_{t-1}|x_t)}}\right]\\ 
&+\mathbb{E}_p\left[
 \log{\frac{p(x_T|x_0)}{p_\theta(x_T)}}\right]\\
 =&\; -\log{p_\theta(x_0|x_1)} + \mbox{\emph{KL}}(p(x_T|x_0) \Vert p_\theta(x_T))\\
&+\sum_{t=2}^{T} \mbox{\emph{KL}}(p(x_{t-1}|x_t,x_0) \Vert p_\theta(x_{t-1}|x_t)).
\end{split}
\end{equation}
\section{Noise estimation.}
\label{noise_estimation}
In this section, we focus on the simplifications suggested by Ho \etal~\cite{ho-NeurIPS-2020} and the adjustments needed to reach the simple objective from Eq.~\eqref{eq_ddpm_4}.

The first simplification is to avoid training the covariance of $p_\theta(x_{t-1}|x_t)$, fixing it beforehand to be $\sigma_t^2\cdot \mathbf{I}$ instead. In practice,  Ho \etal~\cite{ho-NeurIPS-2020} propose to use $\sigma_t^2=\beta_t$. This change impacts the Kullback-Leibler term of $\mathcal{L}_{\scriptsize{\mbox{\emph{vlb}}}}$, because if the covariance is not trainable, then the divergence can be rewritten as being the distance between the means of the distributions plus some constant that does not depend on $\theta$:
\begin{equation}
\label{first_simplification}
\begin{split}
    \mathcal{L}_{\scriptsize{\mbox{\emph{kl}}}} &= \mbox{\emph{KL}}(p(x_{t-1}|x_t,x_0) \Vert p_\theta(x_{t-1}|x_t))\\
    &= \frac{1}{2\cdot\sigma_t^2}\cdot\lVert \Tilde{\mu}(x_t, x_0) - \mu_\theta(x_t,t)\rVert^2 + C,
\end{split}
\end{equation}
where $\Tilde{\mu}(x_t, x_0)$ is the mean of $p(x_{t-1}|x_t,x_0)$, $\mu_\theta(x_t,t)$ is the mean of $p_\theta(x_{t-1}|x_t)$ and C is a constant. We underline that at this point, the output of the neural network is $\mu_\theta(x_t,t)$. 

The next change is based on the observation that the mean $\Tilde{\mu}(x_t, x_0)$ can be expressed as a function of $x_t$ and $z_t$, as follows:
\begin{equation}
\label{mu_tilde}
\begin{split}
    \Tilde{\mu}(x_t, x_0) = \frac{1}{\sqrt{\alpha_t}}\left(x_t - \frac{\beta_t}{\sqrt{1-\hat{\beta_t}}}\cdot z_t\right).
\end{split}
\end{equation}
This implies, according to Eq.~\eqref{first_simplification}, that $\mu_\theta(x_t, t)$ has to approximate this expression. However, $x_t$ is the input of the model. Therefore, Ho \etal~\cite{ho-NeurIPS-2020} propose to reparametrize $\mu_\theta(x_t, t)$ in the same way:
\begin{equation}
\label{mu_theta}
\begin{split}
    \mu_\theta(x_t, t) = \frac{1}{\sqrt{\alpha_t}}\left(x_t - \frac{\beta_t}{\sqrt{1-\hat{\beta_t}}}\cdot z_\theta(x_t,t)\right),
\end{split}
\end{equation}
where $z_\theta(x_t,t)$ is now the output of the neural network, namely an estimation for the noise $z_t$, given the noisy image $x_t$.

If we replace the means in $\mathcal{L}_{\scriptsize{\mbox{\emph{kl}}}}$ with the parametrizations from Eq.~\eqref{mu_tilde} and Eq.~\eqref{mu_theta}, the result is the following:
\begin{equation}
\label{second_simplification}
\begin{split}
    \mathcal{L}_{\scriptsize{\mbox{\emph{kl}}}} = \frac{\beta_t^2}{2\sigma_t^2\alpha_t(1-\hat{\beta}_t)}\lVert z_t -z_\theta(x_t,t)\rVert^2.
\end{split}
\end{equation}
This term is essentially a time-weighted distance between the true noise of the image $x_t$ and the estimation of the network. Ho \etal~\cite{ho-NeurIPS-2020} simplify this term even further and discard the weights $\frac{\beta_t^2}{2\sigma_t^2\alpha_t(1-\hat{\beta}_t)}$, yielding a form that also covers the first term of $\mathcal{L}_{\scriptsize{\mbox{\emph{vlb}}}}$. Therefore, with this latter changes in place, the final objective becomes the simplified version from Eq.~\eqref{eq_ddpm_4}:
\begin{equation}
    \label{loss_simple}
    \mathcal{L}_{\scriptsize{\mbox{\emph{simple}}}} = \mathbb{E}_{t\sim[1,T]}\mathbb{E}_{x_0 \sim p(x_0)}\mathbb{E}_{z_t \sim \mathcal{N}(0,\mathbf{I})}\!\left\lVert z_t-z_\theta(x_t,t)\right\rVert^{2}.
\end{equation}
\end{appendices}
\end{document}